%% file: main.tex
\documentclass[]{xiaomiev}
\microtypesetup{expansion=false} 
\usepackage[page,header]{appendix}
\usepackage{booktabs}
\usepackage{multirow}
\usepackage{graphicx}
\usepackage{tabularx}
\usepackage{hyperref}
\usepackage{amsmath}
\usepackage{amssymb}
\usepackage{makecell} 
\usepackage{pifont}
\usepackage[colorinlistoftodos]{todonotes}
\usepackage{float}
\usepackage{wrapfig}
\setcitestyle{authoryear,round}
\usepackage{enumitem}
\usepackage{listings}
\usepackage[ruled,linesnumbered]{algorithm2e}
\usepackage{tikz}
\usepackage{pgfplots}
\usepackage[table]{xcolor}
\usepackage{colortbl}
\definecolor{miBest}{HTML}{FF7E00}
\definecolor{miSecond}{HTML}{FFBE80}
\pgfplotsset{compat=1.18}
\usetikzlibrary{arrows.meta, positioning, fit, backgrounds, calc, patterns}
\setlist[itemize]{leftmargin=15pt}
\definecolor{hxOrange}{HTML}{ED722E}
\definecolor{hxNavy}{HTML}{1B262C}
\definecolor{hxBlue}{HTML}{0F4C75}
\definecolor{hxTeal}{HTML}{3282B8}
\definecolor{hxSky}{HTML}{BBE1FA}
\definecolor{hxKey}{HTML}{0F4C75}
\definecolor{hxStr}{HTML}{8A5A00}
\definecolor{hxComment}{HTML}{6B7B8C}
\lstdefinelanguage{yaml}{
  sensitive=true,
  morecomment=[l]{\#},
  morestring=[b]",
  morestring=[b]',
  commentstyle=\color{hxComment}\itshape,
  stringstyle=\color{hxStr},
  keywords={true,false,null},
  keywordstyle=\color{hxTeal}\bfseries,
  moredelim=**[il][\color{hxKey}\bfseries]{:},
  moredelim=[l][\color{hxTeal}]{-\ },
}
\newtcblisting{promptfile}[2][]{%
  breakable, listing only,
  colback=hxSky!10, colframe=hxNavy, colbacktitle=hxNavy, coltitle=white,
  boxrule=0.6pt, left=4pt, right=4pt, top=3pt, bottom=3pt,
  title={\scriptsize\textbf{\texttt{#2}}}, #1}
\newtcblisting{manifestfile}[2][]{%
  breakable, listing only,
  listing options={language=yaml},
  colback=hxTeal!8, colframe=hxBlue, colbacktitle=hxBlue, coltitle=white,
  boxrule=0.6pt, left=4pt, right=4pt, top=3pt, bottom=3pt,
  title={\scriptsize\textbf{\texttt{#2}}}, #1}

\newtcolorbox{manifestcard}[1]{%
  colback=hxTeal!8, colframe=hxBlue, colbacktitle=hxBlue, coltitle=white,
  fonttitle=\scriptsize\bfseries, fontupper=\footnotesize,
  boxrule=0.6pt, left=5pt, right=5pt, top=4pt, bottom=5pt,
  before upper={\setlength{\parindent}{0pt}\raggedright},
  title={#1}}
\newtcolorbox{seedbox}[2][]{%
  enhanced, breakable,
  colback=hxSky!5, colframe=hxBlue!65, colbacktitle=hxSky!18, coltitle=hxBlue,
  fonttitle=\small\bfseries, fontupper=\small,
  boxrule=0.35pt, leftrule=1.6pt, arc=1mm,
  left=5pt, right=5pt, top=4pt, bottom=4pt,
  before skip=6pt, after skip=6pt,
  before upper={\setlength{\parindent}{0pt}\raggedright},
  title={#2}, #1}

\newlength{\ataglancegap}
\newtcolorbox{seednote}[2][]{%
  enhanced, breakable,
  colback=hxOrange!5, colframe=hxOrange!70, colbacktitle=hxOrange!16, coltitle=hxStr,
  fonttitle=\small\bfseries, fontupper=\small,
  boxrule=0.35pt, leftrule=1.6pt, arc=1mm,
  left=5pt, right=5pt, top=4pt, bottom=4pt,
  before skip=6pt, after skip=6pt,
  before upper={\setlength{\parindent}{0pt}\raggedright},
  title={#2}, #1}

\RequirePackage{xspace}
\makeatletter
\DeclareRobustCommand\onedot{\futurelet\@let@token\@onedot}
\def\@onedot{\ifx\@let@token.\else.\null\fi\xspace}

\makeatother

\newcommand{\CodePhysical}{\textcolor{hxOrange}{HarnessPAI}\xspace}

\title{{\fontsize{14.5}{18}\selectfont \textcolor{hxOrange}{HarnessPAI}: An Evolving Harness for Physical AI}}

\author{\textbf{Darwin Agent Team}\\
{\small See \hyperref[sec:contributions-acknowledgments]{Contributions and Acknowledgments} section for a full author list.}\\[2mm]
\includegraphics[
  width=\linewidth,
  height=18mm,
  keepaspectratio
]{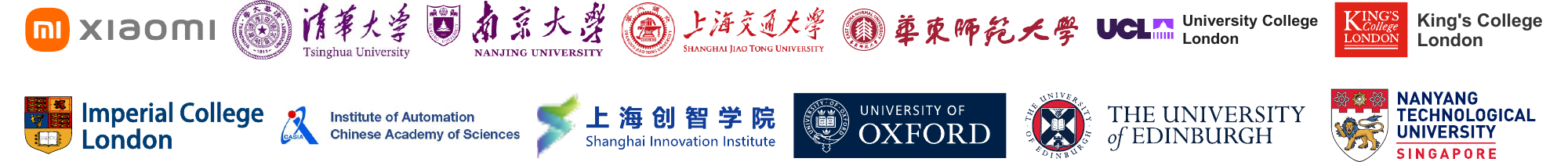}}
\renewcommand{\abstractinfont}{\fontsize{9}{10.5}\selectfont}
\abstract{
Physical AI aims to build embodied agents that \emph{perceive} the world, \emph{understand and reason} about it, and decide how to \emph{act}.
Yet the field has focused primarily on the last component: the \emph{action model} that maps observations to low-level controls.
The prevailing training recipe can erode the perceptual and reasoning capabilities needed for robust behavior, leaving even strong action models vulnerable to scene perturbations and long-horizon tasks.
We introduce \textbf{\CodePhysical{}}, a model- and embodiment-agnostic \textbf{Harness} framework for \textbf{P}hysical \textbf{AI} that treats code as the executable and evolvable interface that organizes the underlying action primitive.
The framework separates two timescales: within a rollout, it executes \emph{open-loop} at the program level, with a fixed program guiding and checking execution; across rollouts, it evolves \emph{closed-loop}, using execution feedback to revise the program and distill failures into reusable skills.
Across desktop robot arms, household robots, a robot vacuum, and a legged walking agent, \CodePhysical{} demonstrates consistent improvements against baselines.
Beyond the harness, the converged program also acts as a low-cost, reliable source of expert data: it generates training data for fine-tuning VLA and WAM models on tasks where it is hard for these models to obtain successful trajectories on their own.
Our results suggest that the frontier of Physical AI depends not only on stronger action models, but also on executable harnesses that integrate perception, task understanding and reasoning, and action execution into a unified, verifiable, and feedback-driven system.

Website: \textcolor{hxOrange}{https://darwin-agent.github.io/HarnessPAI}.
}

\begin{document}
\maketitle

\begin{figure}[H]

\centering
\includegraphics[width=\linewidth]{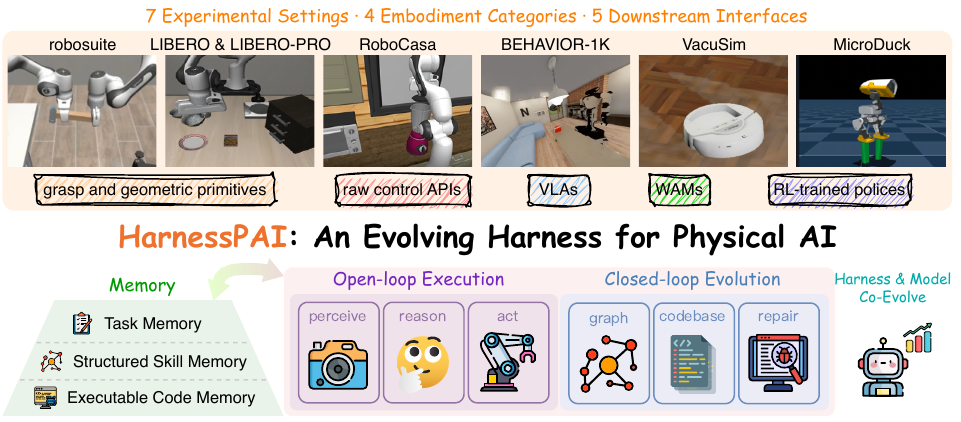}
\caption{
Overview of \CodePhysical{}, a general physical harness framework. }
\label{fig:overview}
\end{figure}

\clearpage
\setcounter{tocdepth}{2}
\tableofcontents
\setcounter{tocdepth}{2}
\clearpage
\input{chapters/1_introduction}
\input{chapters/2_related_work}
\input{chapters/3_motivation}
\input{chapters/4_framework}
\input{chapters/4_skill_library}
\input{chapters/4_model_feedback}
\input{chapters/5_experiment}
\input{chapters/6_discussion}
\input{chapters/7_conclusion}
\FloatBarrier
\bibliographystyle{plainnat}
\clearpage
\bibliography{references}
\clearpage
\input{chapters/acknowledgments}
\clearpage
\input{chapters/appendix}

\end{document}

%% file: chapters/1_introduction.tex
\section{Introduction}
\label{sec:introduction}
Physical AI seeks to build embodied agents that can \emph{perceive} their surroundings, \emph{understand and reason} about their goals, and \emph{act} effectively in the physical world~\citep{harnessvla,aspire,capx,ding2026zetta,openeta}.
Recent foundation models have substantially advanced the first two capabilities, but closing the loop through action remains the hardest step, because physical behavior must satisfy constraints imposed by geometry, dynamics, contact, and safety over extended horizons and under uncertainty.
This challenge has made learned action models a central focus of Physical AI~\citep{pi05,openvla,ye2026dreamzero}.

Two complementary architectures have emerged as prominent approaches to learned action.
Vision-Language-Action models (VLAs) map visual observations and language instructions directly to robot actions~\citep{rt1,rt2,pi0,pi05,openvla,grootn1}, while World Action Models (WAMs) additionally model how the environment evolves under those actions~\citep{worldvla,ye2026dreamzero}.
Their complementary strengths point toward a unified architecture that combines semantic grounding, world prediction, and action generation.
Strategies for adapting these models to downstream physical tasks broadly follow two directions.
Post-training updates model parameters using additional demonstrations or interaction.
However, fine-tuning can narrow the learned action distribution and hurt generalization, and its gains remain bounded by the base model itself.
Their performance remains constrained by the scale, diversity, and quality of available embodied data, as well as by the difficulty of optimizing semantic understanding and physical control jointly.
Indeed, even VLAs that approach perfect success on standard benchmarks can fall sharply under controlled scene and task perturbations~\citep{liberopro}.
Harness-based approaches instead keep the action model fixed and improve the system around it through external perception, reasoning, orchestration, and verification~\citep{harnessvla,roboharness}.
We take the latter view, treating the learned model as an action primitive organized and complemented by an executable harness.

In language and multimodal AI, harnesses have turned foundation models into capable agents by organizing tools, memory, reasoning, and iterative execution~\citep{chen2026harnessx,liu2026mimemory,qiao2026memoryintelligenceagent,wu2026trace}.
Recent approaches such as Harness VLA and RoboHarness extend this systems perspective to Physical AI, retaining a frozen action model as a low-level primitive and placing memory, task decomposition, skill selection, and failure recovery around it, which preserves learned motor competence while making high-level behavior more adaptive~\citep{harnessvla,roboharness}.
However, current physical harnesses still rely heavily on online reasoning and memory accumulation: an agent must repeatedly interpret observations, retrieve experience, and decide what to do, even when executing the same task again.
Executable code offers a natural representation for consolidating and reusing this system-level knowledge, yet code-as-policy approaches replace learned manipulation with hand-designed APIs, sacrificing contact-rich control~\citep{codeaspolicies,aspire,capx}.
What is missing is an executable harness that retains the learned action model, compiles perception, reasoning, action-model invocation, and verification into reusable programs, and evolves those programs from physical feedback across rollouts.

We introduce \textbf{\CodePhysical{}}, a model- and embodiment-agnostic harness framework for Physical AI that treats code as the executable and evolvable interface organizing the underlying action primitive.
The framework separates execution and improvement into two complementary timescales.
Within each rollout, it performs \emph{open-loop execution}: an agent instantiates an executable program fixing the task decomposition, perception calls, geometric reasoning, action-model invocations, and success checks, which then runs directly without further high-level deliberation, delegating contact-rich manipulation to the underlying action model.
Across rollouts, it performs \emph{closed-loop evolution}: observations, execution traces, failures, and outcomes are collected as physical feedback to diagnose errors, revise the program, and distill reusable failure-to-repair into a graph structured skill memory.
Code is therefore both an action interface and an evolvable representation of agent behavior---readable enough to diagnose, modular enough to edit, and executable enough to ground each revision in the physical environment.
We further show that the evolved program attains high reliability and quality on tasks of the same type, and serves as a reliable expert-data collector whose successful trajectories improve the wrapped action model itself.

Compared with recent physical harnesses, \CodePhysical{} changes the unit of adaptation from transient online reasoning to persistent executable behavior.
Existing systems often retrieve memories, replan, or intervene through repeated agent calls during deployment.
\CodePhysical{} instead compiles successful reasoning into code, validates it across scenes, and reuses the resulting program when the task recurs.
This removes online deliberation from the critical path and amortizes its cost across executions, while preserving the contact-rich motor competence of learned action models.

Rapid advances in general-purpose agents make this distinction increasingly important.
A frontier model may reason, write code, and invoke available tools, but its success still depends on those tools being grounded in physical perception, embodiment-specific control, and reliable feedback.
Physical actions are continuous, contact-rich, and often difficult to reverse, requiring an execution substrate that is fast, reproducible, and explicitly verifiable.
\CodePhysical{} provides this substrate.
Stronger models can improve code generation and diagnosis during evolution, but they do not replace the validated programs, physical primitives, and accumulated repair knowledge needed at deployment.
\CodePhysical{} therefore complements advances in foundation models by turning transient model reasoning into persistent physical capability.

We evaluate \CodePhysical{} in seven experimental settings spanning four broad embodiment categories---robot arms, a humanoid upper body on a mobile base, a mobile robot vacuum, and a legged walking agent.
It improves substantially on the action model it wraps in every one of them.
On LIBERO-PRO, where policies above 90\% on standard LIBERO fall sharply under perturbation, it holds 96.5\%, 8.8 points above the strongest harness baseline.
Fine-tuning $\pi_{0.5}$ on the self-collected trajectories lifts LIBERO-PRO success from 34.9\% to 73.7\%: the harness does not only improve the system around the model, it also improves the model itself.
The same approach lifts DreamZero's mean success on LIBERO-Object Swap and Task from 5.1\% to 26.5\%, extending this benefit to a WAM.
Our core contributions are:
\begin{itemize}
    \item \textbf{Executable harness abstraction.} We introduce HarnessPAI, a model- and embodiment-agnostic framework that organizes perception, task-level reasoning, verification, and learned physical actions through executable programs.
    \item \textbf{Two-timescale evolution.} We develop a two-timescale execution and evolution procedure: programs execute without online high-level revision within a rollout and are revised across rollouts using execution feedback.
    \item \textbf{Structured skill memory.} We introduce a node-centric skill memory that stores failure-to-repair knowledge and transfers it across tasks sharing the same workflow stages.
    \item \textbf{Empirical validation and model feedback.} We evaluate the framework across heterogeneous embodiments and action backends, and show that converged harness trajectories can improve the wrapped action model through post-training.
\end{itemize}

%% file: chapters/2_related_work.tex
\section{Related Work}
\label{sec:related}

Physical AI aims to build embodied agents that \emph{perceive} the world, \emph{understand and reason} about it, and decide how to \emph{act}.
Foundation models have largely supplied the first two capabilities, while effort has concentrated on the third: the \emph{action model} that maps observations to low-level controls, and where the competence to act should live. Existing approaches differ primarily in where adaptation is stored: in model parameters, in executable programs, or in an external runtime harness.

\subsection{Post-training VLAs and WAMs}
\label{sec:related-post-training}

A growing body of work brings the success of LLMs and VLMs to embodied control, yielding a diverse family of VLAs such as RT-1/RT-2~\citep{rt1,rt2}, OpenVLA~\citep{openvla}, the $\pi$-series~\citep{pi0,pi05,pi2025pi06,pi2026pi07}, GR00T~\citep{grootn1}, and Octo~\citep{octo}, which map visual observations and language instructions to robot actions.
These models span several action-generation designs: OpenVLA autoregressively predicts discretized action tokens with an open 7B VLM backbone~\citep{openvla}; $\pi_0$ and $\pi_{0.5}$ condition a flow-matching action expert on a pretrained vision-language backbone, with $\pi_{0.5}$ extending the setting toward open-world and long-horizon manipulation~\citep{pi0,pi05}; and GR00T N1 combines a vision-language reasoning module with a diffusion-transformer controller for generalist humanoid robotics~\citep{grootn1}.
Despite this progress, VLAs remain markedly more fragile than their LLM/VLM counterparts, frequently failing on tasks that lie only modestly outside their training distribution~\citep{liberopro,xie2026sva}.
A complementary family of WAMs couples action generation with prediction of future visual states, using video prediction to learn the consequences of actions and to support long-horizon control~\citep{worldvla,ye2026dreamzero,yuan2026fastwam,wu2024gr1,cheang2024gr2,zhu2025uwm}.
WorldVLA jointly models action and future observations in an autoregressive action-observation formulation~\citep{worldvla}; DreamZero scales this idea with a pretrained video-diffusion backbone and jointly predicts visual futures and actions from heterogeneous robot data~\citep{ye2026dreamzero}; and Fast-WAM shows that much of the benefit can come from video modeling during training, without rendering future frames at test time~\citep{yuan2026fastwam}.
WAMs therefore enrich the action model with predictive physical structure, but their adaptation still occurs through training the model and its associated objectives.
\CodePhysical{} is orthogonal to these model-side approaches: it keeps the action model fixed during harness evolution and stores task-specific adaptation in executable programs and reusable skills rather than in model parameters.

\subsection{Code as Policy}
\label{sec:related-code-as-policy}

In code-as-policy systems, the generated program serves as the primary task policy: it translates a natural-language instruction into a sequence of calls to perception, geometry, motion, or simulation APIs~\citep{huang2023voxposer,codeaspolicies,singh2023progprompt,capx,aspire}.
This representation makes high-level behavior compositional and inspectable: a generated program can sequence primitives, branch on sensed state, and invoke reusable skills instead of emitting an undifferentiated action sequence.
In the original code-as-policy formulation, language models generate Python policies that combine visual affordances with a library of low-level robot interfaces~\citep{codeaspolicies}.
Subsequent work has explored more structured skill composition and agentic program synthesis.
CaP-X studies coding agents for robot manipulation as a benchmark and improvement problem, making the quality of generated code, tool use, and iterative debugging explicit evaluation axes~\citep{capx}.
ASPIRE discovers and reuses robot skills through an agent that writes, executes, and self-corrects programs against an explicit skill library~\citep{aspire}.
However, these methods usually obtain low-level action competence from hand-designed primitives or direct control APIs. Such interfaces can provide precise geometric control, but they may lack the contact-rich robustness and learned adaptability of pretrained action models, especially when object contacts, intermediate states, or scene conditions differ from the assumed execution pattern.
\CodePhysical{}, in contrast, uses code to organize, invoke, and verify a learned action model rather than relying solely on control APIs, preserving its physical competence while moving reusable perception, reasoning, and task logic into an evolvable executable harness with structured skill memory.

\subsection{Harnesses for VLAs and WAMs}
\label{sec:related-harnesses}

Harness methods preserve pretrained physical competence while adding an external layer for planning, memory, verification, and recovery~\citep{harnessvla,chen2026showharness,xu2026baton,liu2026phyagentos}.
Harness VLA steers a frozen VLA through memory-guided task decomposition and skill selection, while Zetta likewise keeps the base policy frozen but self-evolves code-based runtime critics and recovery skills through action-, rollout-, and update-timescale loops~\citep{harnessvla,ding2026zetta}.
Show-Harness instead exposes semantic action units that a VLM selects directly and embodiment-specific interpreters deterministically ground into robot motion, enabling one interface across tasks and embodiments~\citep{chen2026showharness}.
RoboHarness further combines experience retrieval, failure attribution, external intervention, and memory consolidation for in-context adaptation~\citep{roboharness}.
Extending harnesses to WAMs, HarnessWAM maintains an evidence-grounded scene belief and structured task graph to organize finite-horizon WAM skills, verify progress, and recover from failures in long-horizon tasks~\citep{gu2026harnesswam}.
Depending on the method, these systems may invoke online reasoning, memory retrieval, semantic action selection, or runtime critic–recovery loops during deployment. This can improve adaptability, but it also places high-level model inference on the execution path.
\CodePhysical{} makes the complete task program the unit of adaptation: a validated program executes directly when the task recurs, while code evolution for new tasks retrieves reusable skills from structured skill memory and can transfer across VLA, WAM, and other action backends.

\subsection{Sim-to-Real Transfer and Real-Robot Validation}
\label{sec:related-sim2real}

Sim-to-real work addresses the reality gap through visual and dynamics randomization, real-data simulator adaptation, and real-to-sim-to-real calibration~\citep{tobin2017domain,peng2018dynamics,chebotar2019simopt,torne2024rialto}; recent methods also use language models for transfer design and online correction, while SIMPLER evaluates sim-real correspondence~\citep{ma2024dreureka,jiang2024transic,li2024simpler}. Unlike these policy-focused approaches, \CodePhysical{} must transfer executable assumptions about perception, geometry, verification, timing, and recovery under sensing error, latency, contact variation, and hardware faults, motivating safety-gated validation on real robots.

\begin{figure}[H]
\centering
\includegraphics[width=\linewidth]{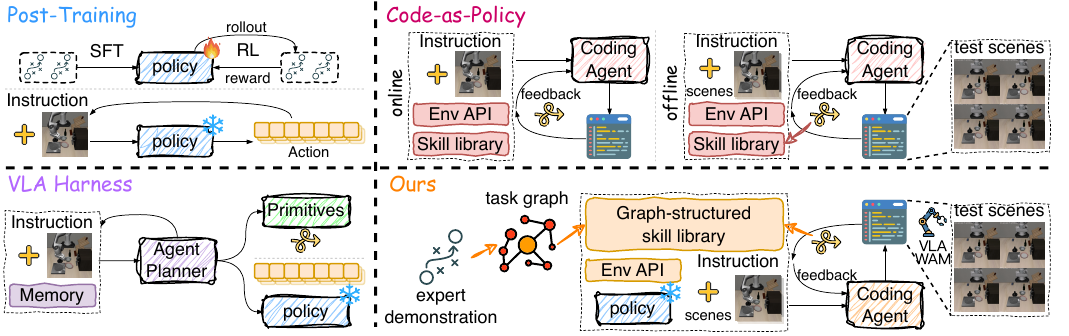}
\caption{Our paradigm versus prior ones. (1) \emph{Post-training}: an
action model is trained on expert trajectories by SFT or by RL on a reward; at inference, the frozen model takes an instruction and image and outputs an action chunk.
(2) \emph{Code as policy}: an \emph{online} variant evolves one program per task and seed, where the coding agent reads the instruction and image, calls APIs and the skill library, runs the code, revises it from the output, and evolves again; an \emph{offline} variant instead evolves a single program over several selected seeds and runs that same program directly on scenes unseen during evolution.
(3) \emph{VLA harness}: a coding agent serves as the planner, receiving the image, instruction, and memory and deciding whether to call an environment primitive or the action-model policy.
Our \CodePhysical{} instead first builds a task graph from expert trajectories, derives the code and skill-library architecture from that graph, and uses a coding agent as the planner that invokes VLA/WAM actions \emph{through code}, so code manages perception, understanding and reasoning, and action execution; for one or many tasks it evolves code over multiple environment seeds and tests a single program on many unseen test scenes.}
\label{fig:related-comparison}
\end{figure}

Figure~\ref{fig:related-comparison} gives a visual overview of these paradigms; Table~\ref{tab:related-comparison} summarizes how they differ in where adaptation resides and how execution is organized.

\begin{table*}[t]
\centering
\small
\setlength{\tabcolsep}{4pt}
\renewcommand{\arraystretch}{1.15}
\renewcommand{\tabularxcolumn}[1]{m{#1}}
\begin{tabularx}{\textwidth}{@{}>{\raggedright\arraybackslash\bfseries}m{2.7cm}>{\raggedright\arraybackslash}m{2.8cm}>{\centering\arraybackslash}m{2.5cm}>{\centering\arraybackslash}m{1.8cm}>{\centering\arraybackslash}m{1.8cm}>{\raggedright\arraybackslash}X@{}}
\toprule
Method family & Adaptation location & Coding agent during rollout & API-based control & Learned policy usage & Evolution across rollouts \\
\midrule
VLA/WAM post-training & Model parameter & \ding{55} & \ding{55} & \ding{51} & Training only \\
Code-as-policy & Program & \ding{55} & \ding{51} & \ding{55} & Sometimes \\
VLA harness & Runtime/memory & \ding{51} & \ding{51} & \ding{51} & Usually \\
HarnessPAI & Program/memory & \ding{55} & \ding{51} & \ding{51} & Designed for persistent evolution \\
\bottomrule
\end{tabularx}
\caption{Qualitative comparison of action-model and executable-harness paradigms, corresponding to Fig.~\ref{fig:related-comparison}.}
\label{tab:related-comparison}
\end{table*}

%% file: chapters/3_motivation.tex
\section{Motivation}
\label{sec:problem-setup}

The introduction argued that a large fraction of Physical AI research is organized around a single component---the \emph{action model}.
Before presenting \CodePhysical{}, we make three questions precise and show why the action-model-centric design answers each of them poorly: why a VLA or WAM is not, on its own, a sufficient physical agent (Section~\ref{sec:why-harness}); why a physical agent should execute \emph{open-loop} at the program level and aim to succeed on the first attempt (Section~\ref{sec:why-openloop}); and why \emph{code} is the right material in which to organize the missing capabilities (Section~\ref{sec:why-code}).

\subsection{Why Do VLA/WAM Models Need a Harness?}
\label{sec:why-harness}

Almost all recent VLA and WAM models are assembled the same way: a pretrained vision-language model (VLM) supplies semantic perception, and a separately trained \emph{action head}---typically a diffusion, flow-matching, or diffusion-transformer model---maps the VLM's representations into continuous action chunks or latent actions.
The $\pi_0$ family learns a flow-matching action head on top of a PaliGemma vision-language backbone~\citep{pi0,pi05}; GR00T N1 conditions a diffusion-transformer action head on vision-language tokens~\citep{grootn1}; Octo fits a diffusion action head over a transformer backbone~\citep{octo}; and WorldDreamer learns a World Action Model that jointly predicts future scene evolution and action chunks~\citep{worlddreamer}.
Even the autoregressive alternatives---RT-2 and OpenVLA---follow the same division, casting actions as additional output tokens of a VLM~\citep{rt2,openvla}.
Regardless of whether actions are emitted through diffusion denoising, flow matching, or discretized tokens.
Despite architectural differences, these systems commonly expose an action-centric interface, with most downstream optimization focused on predicting or evaluating physical actions.

\begin{figure}[htbp]
\centering
\includegraphics[width=\linewidth]{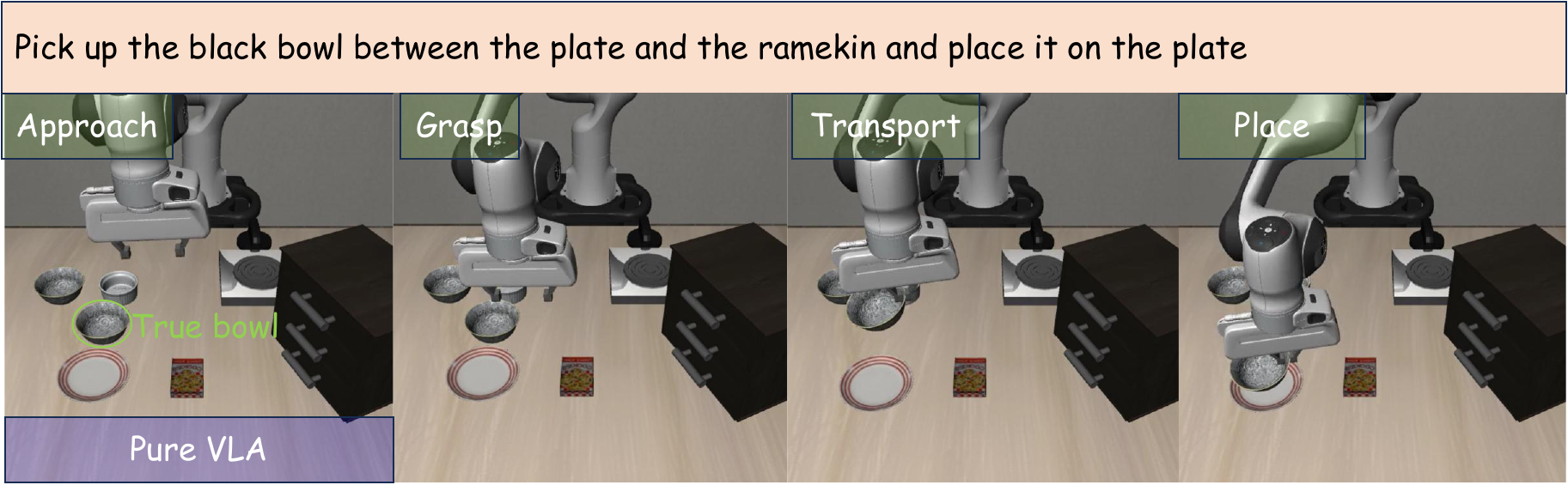}
\caption{In-distribution success of a strong VLA ($\pi_{0.5}$) on LIBERO. Given
the in-distribution instruction ``pick up the black bowl between the plate and the ramekin,'' the VLA approaches, grasps the correct bowl, transports it, and places it on the plate.}
\label{fig:motivation-success}
\end{figure}

\begin{figure}[htbp]
\centering
\includegraphics[width=\linewidth]{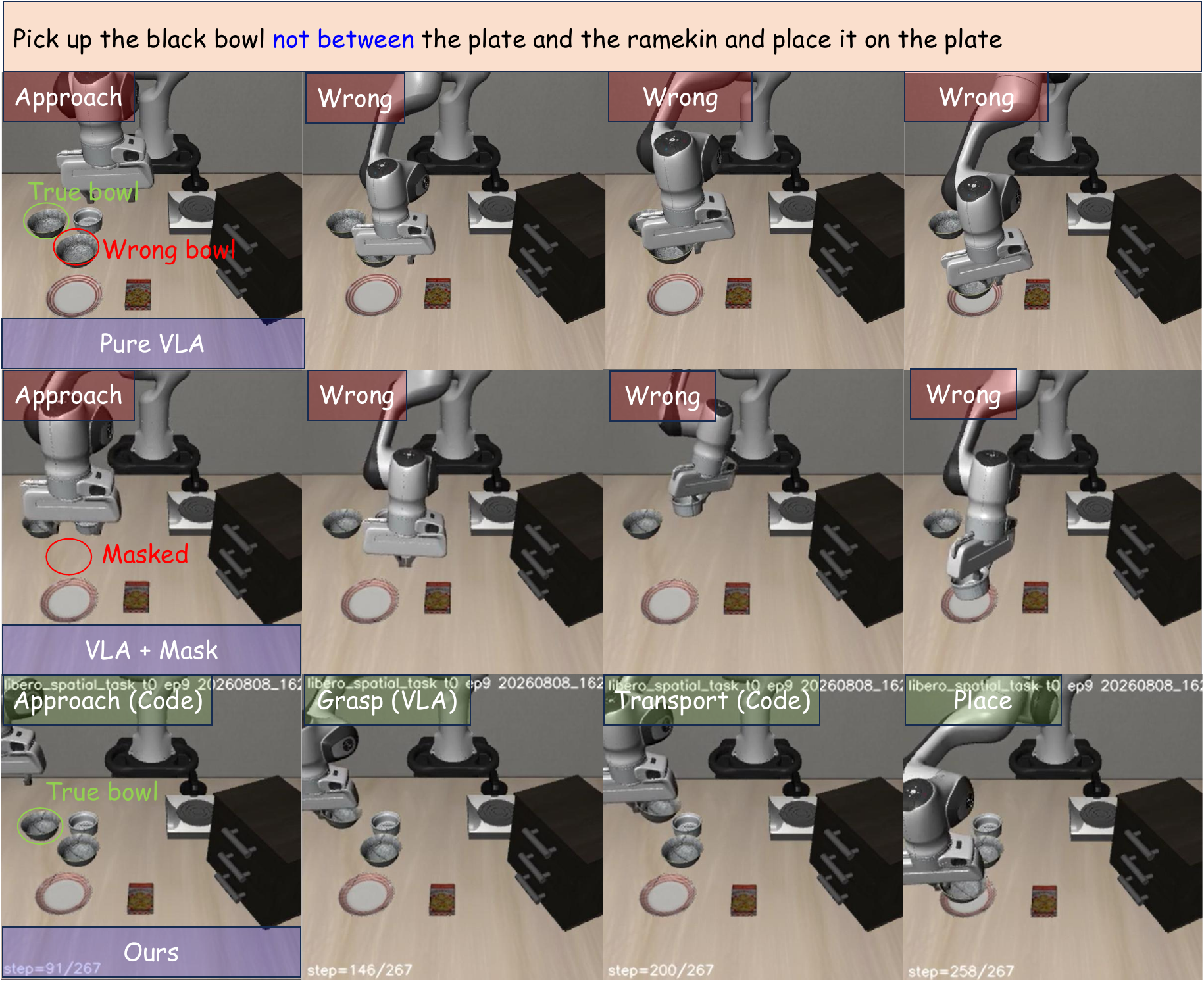}
\caption{Language perturbation on LIBERO. Adding ``not'' before ``between''
flips the intended target to a different bowl, yet $\pi_{0.5}$ still grasps the original bowl, and externally masking the scene does not help.
\CodePhysical{} re-grounds the instruction in code and grasps the correct bowl.}
\label{fig:motivation-language}
\end{figure}

\begin{figure}[htbp]
\centering
\includegraphics[width=\linewidth]{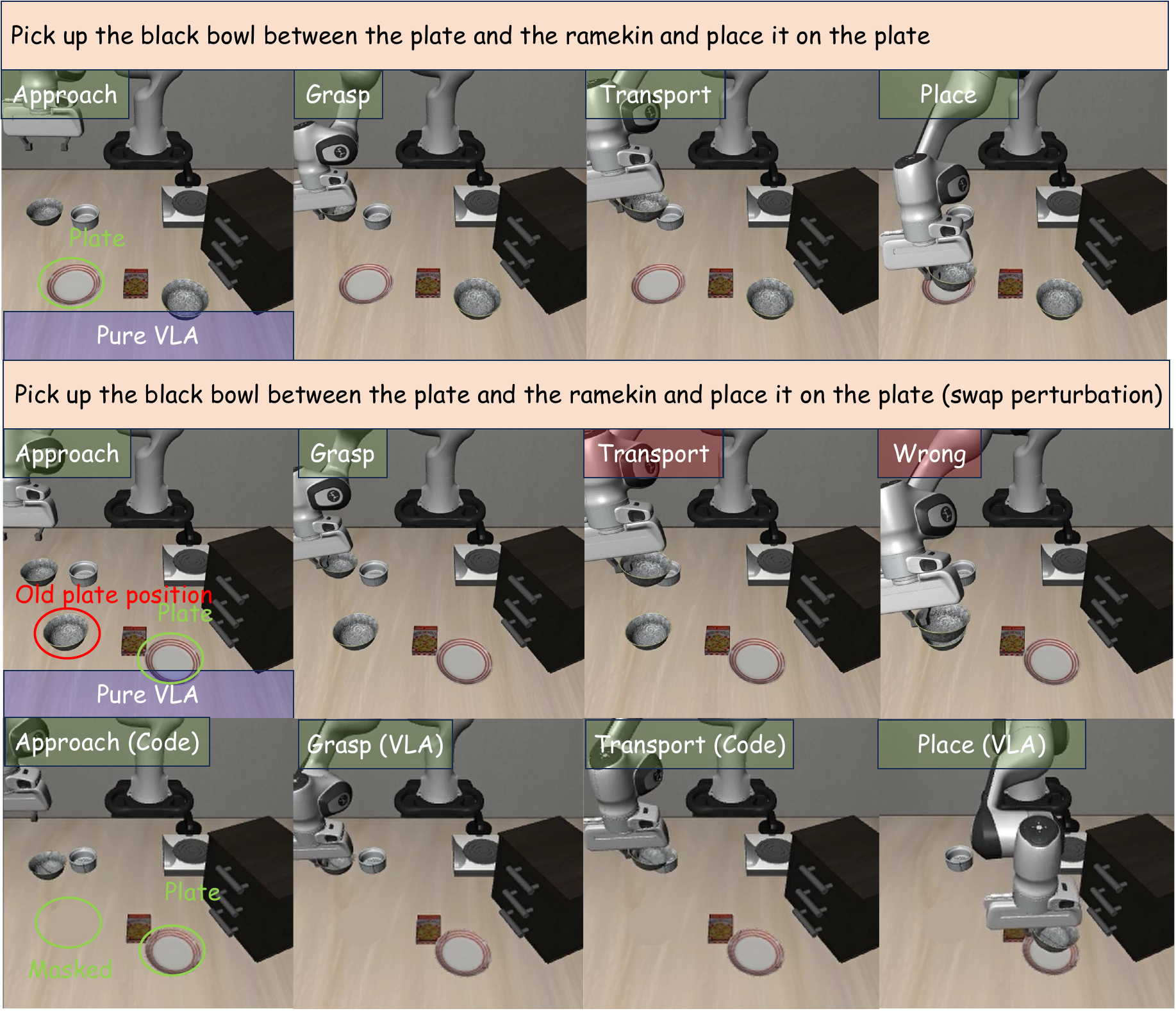}
\caption{Visual perturbation on LIBERO. After the positions of the plate and
ramekin are swapped without changing the instruction, $\pi_{0.5}$ grasps the bowl but attempts to place it at the plate's original location and fails.
\CodePhysical{} re-locates the objects in code and succeeds.}
\label{fig:motivation-vision}
\end{figure}

Action models learn demonstrated behavior through behavior cloning and can also be optimized for task rewards through reinforcement learning.
However, strong action-task performance does not guarantee reliable target grounding and task completion when the instruction or scene changes.
The following examples illustrate this limitation, but do not establish whether it results from degradation of vision and language capabilities during training.
Fig.~\ref{fig:motivation-success} shows $\pi_{0.5}$ succeeding in an in-distribution scene, while Figs.~\ref{fig:motivation-language} and~\ref{fig:motivation-vision} show it failing under a language and a visual perturbation, respectively.

Fig.~\ref{fig:motivation-language} illustrates a language perturbation.
The demonstrated task is ``pick up the black bowl between the plate and the ramekin.''
Adding ``not'' before ``between'' flips the intended referent to a \emph{different} bowl.
A model that understood the instruction would re-derive the referent and grasp the correct object.
$\pi_{0.5}$ does not: it still approaches and grasps the bowl between the plate and the ramekin despite the added negation.
The ``VLA + Mask'' row shows that the illustrated masking intervention still does not enable the model to execute the modified instruction correctly.
This particular masking intervention does not resolve the failure, suggesting that simply modifying the visual input is insufficient in this case.
This indicates that the visual intervention is insufficient to resolve the instruction-following error, but does not by itself determine whether the error arises from language understanding, visual grounding, or the mapping from instructions to actions.

Fig.~\ref{fig:motivation-vision} illustrates a visual perturbation.
Here the instruction is unchanged, but the positions of the plate and the ramekin are swapped.
A visually grounded agent would re-locate the referenced objects and adjust its motion accordingly.
Instead, in the illustrated rollout, $\pi_{0.5}$ grasps and transports the bowl but attempts to place it at the plate's \emph{original} location rather than its current location.
The model completes the grasp and transport but fails to update the placement target to reflect the changed layout.
This is exactly the collapse that LIBERO-PRO exposes---models above 90\% on standard LIBERO lose most of their accuracy under such scene perturbations~\citep{liberopro}.

\begin{figure}[htbp]
\centering
\includegraphics[width=1\linewidth]{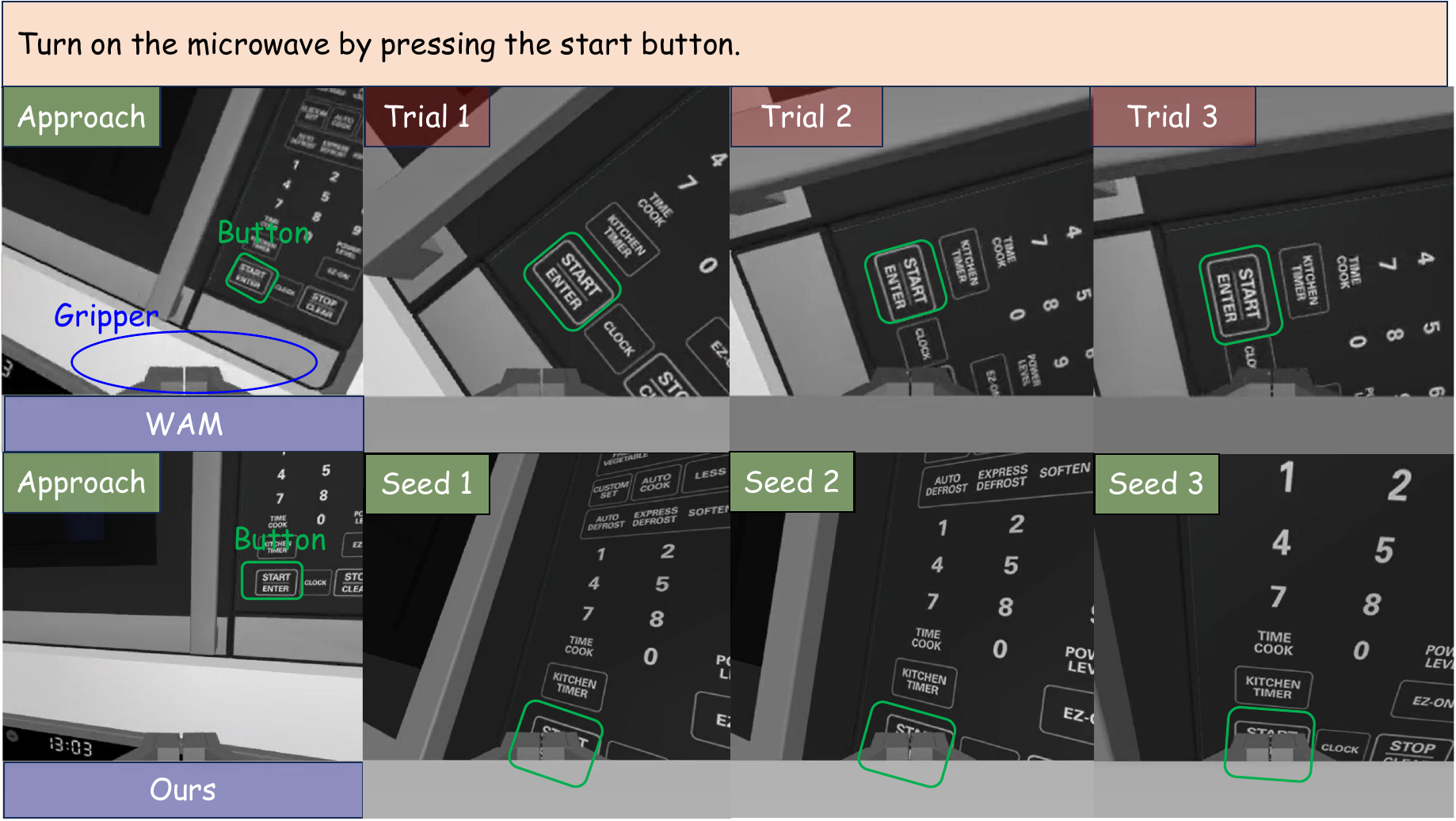}
\caption{Fine-manipulation error on RoboCasa. The evaluated WAM reaches the microwave but
fails to press the small start button in the illustrated trials.
\CodePhysical{} successfully completes the button-pressing task.}
\label{fig:motivation-precision}
\end{figure}

Even when the robot reaches the vicinity of the target, a further failure mode appears in fine manipulation.
The diffusion and flow-matching action heads that most modern policies use are generative models over continuous action chunks.
Fig.~\ref{fig:motivation-precision} shows a button-pressing task: the evaluated WAM approaches the microwave and reaches toward the start button, but fails to depress it in the illustrated trials.
These results alone do not establish whether the failure arises primarily from target localization, contact control, or action generation.

Taken together, these observations answer the first question.
These examples show that action competence alone does not guarantee reliable task completion under instruction changes, scene perturbations, or fine-manipulation demands.
We therefore investigate an external harness that complements the action model with explicit target grounding, geometric reasoning, and execution checks, while keeping the underlying model fixed.

\subsection{Why Open-Loop Execution?}
\label{sec:why-openloop}

\begin{figure}[htbp]
\centering
\includegraphics[width=\linewidth]{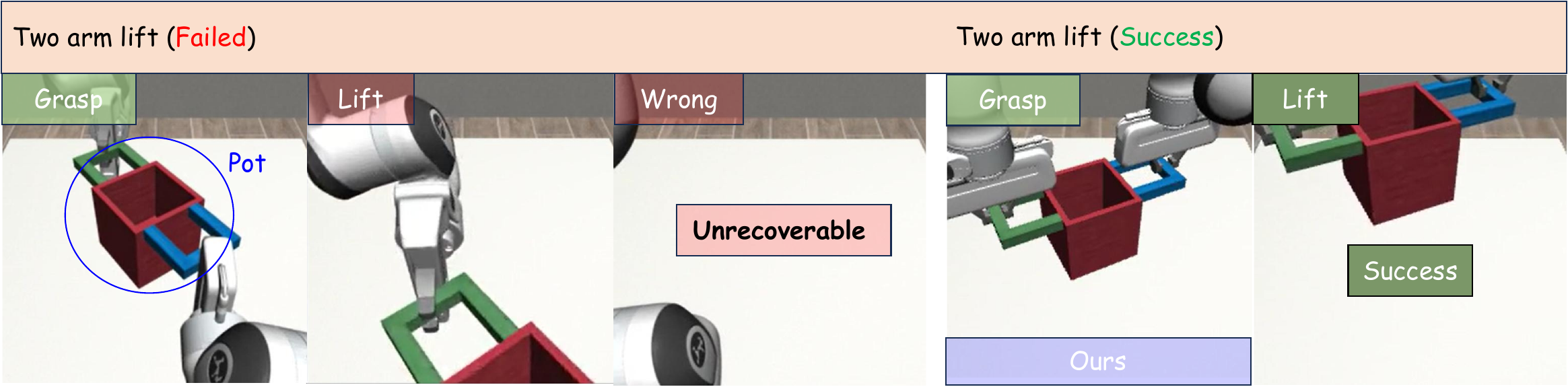}
\caption{Irreversible mistake on RoboCasa.  The robotic arm may knock over the pot, and this mistake cannot be undone, ending the rollout in an unrecoverable state.
\CodePhysical{} grounds the grasp in code and succeeds on the first attempt.}
\label{fig:motivation-irreversible}
\end{figure}

A second problem shapes the design of any such harness: physical mistakes are frequently irreversible.
Fig.~\ref{fig:motivation-irreversible} shows a two-arm lift in which the WAM grasps the wrong object and lifts it.
In the illustrated rollout, the incorrect grasp changes the scene state, and the subsequent execution fails to recover and complete the task.
Unlike cases where execution can be retried after resetting the program state, physical actions can leave persistent changes in the environment: some errors require additional recovery actions, while others may have irreversible consequences.
A physical agent should therefore seek to identify potential errors before executing high-risk actions, with the aim of improving first-attempt success and reducing the cost of failure.

These risks motivate organizing task execution within a stable, inspectable program.
Here, \emph{open-loop execution} refers specifically to the program level: the harness program remains fixed within each rollout, while its execution can still use current observations and predefined checks to ground targets, carry out actions, and determine whether to proceed to the next step.
Program revisions occur between rollouts.
We revise the program using execution feedback between rollouts and evaluate the revised program in experimental environments that permit resets.
Within a rollout, execution may still use predefined checks and recovery mechanisms; the availability of environment resets should not be assumed in real-world deployment.
These two mechanisms are complementary: explicit grounding and checks within a rollout aim to reduce the risk of incorrect actions, while evolution across rollouts uses execution feedback to improve the program and the reliability of subsequent executions.

\subsection{Why Use Code as a Physical AI Harness?}
\label{sec:why-code}

The first two questions point to a specific locus of improvement.
Rather than repeatedly fine-tuning the action model in the hope of improving its visual grounding, language understanding, and manipulation precision, we can support these capabilities through an external harness that executes open-loop and evolves closed-loop.
This leaves a final question: why express that harness as \emph{code}?

Code is the natural material for three reasons.
First, it is \emph{executable}: a program grounds each task in the current scene, decomposes it into steps, invokes the action model as a primitive, and runs explicit success checks---so open-loop execution is, in effect, running a program that checks task success as it runs.
Second, it is \emph{inspectable}: perception, grounding, and recovery decisions are written as statements that can be read, diagnosed, and edited, rather than hidden inside a monolithic policy's weights.
Third, it is \emph{evolvable}: a program can be revised in response to feedback and then re-executed, which is precisely the closed-loop evolution the first two questions require.
Code can wrap different VLA and WAM backbones as interchangeable primitives, and degenerates to direct API control when no learned action model is available.

One might still object that a program only solves the task it was written for: if \CodePhysical{} merely instantiates a hand-written program, it would seem to handle only the tasks whose code already exists.
This objection rests on a conflation of two deployment regimes.

\begin{seednote}{Key question}
In a real deployment, does the agent face \emph{many tasks, each executed once}, or \emph{one task, executed many times}?
\end{seednote}

A pure VLA or WAM is built for the first regime, with the goal of generating correct actions at inference time from previously unseen instructions.
\CodePhysical{}, by contrast, targets the second regime, in which the same physical task is executed repeatedly.
But executing the same task repeatedly does not relax the requirements for each individual run.
Each run remains \emph{open-loop} at the program level; because physical errors may be difficult to recover from or even have irreversible consequences, the agent should still aim to complete the task correctly on the first attempt.
The many-times regime is therefore not a license to tolerate sloppy attempts; instead, it provides opportunities for closed-loop evolution across rollouts to accumulate feedback and refine the program before the next run, with the aim of improving the reliability of subsequent executions.

Moreover, the repairs discovered for one task do not stay with that task.
\CodePhysical{} accumulates repairs distilled from previous tasks into a \emph{how-to-repair} skill library for reuse when a new task arrives.
Examples include re-localizing the target after a missed grasp, verifying a press before moving on, and re-grasping after a slip.
A novel task is therefore not solved from scratch: its first failure is matched against previously learned repair skills, so the framework generalizes quickly to new scenes and instructions even though it has never seen their code before.
Executability, inspectability, and evolvability, together with a how-to-repair skill library that transfers across tasks, are the properties that \CodePhysical{}, described next, formalizes.

%% file: chapters/4_framework.tex
\section{Framework}
\label{sec:framework}

\subsection{Open-Loop Execution}
\label{sec:openloop}

Because physical actions can be costly or difficult to reverse, \CodePhysical{} uses a fixed program with explicit checks to reduce execution errors within each rollout.
To support this goal, the program coordinates perception of task-relevant objects, geometric reasoning, and action execution.
The examples in Section~\ref{sec:why-harness} show that successful action generation does not guarantee reliable grounding or instruction following under perturbations, motivating explicit support for these capabilities.
\CodePhysical{}'s response is structural rather than parametric: it keeps the three capabilities \emph{separate} and \emph{explicit}, and organizes them into a program that checks itself as it runs.
Verification here is not a retry loop that paws at the world until it finally sticks; it is a \emph{success test}--- has the grasp actually closed on the object, is the object where it should be--- that \emph{gates the next step}, so the agent commits to \emph{transport} only once the grasp has been confirmed.
\begin{figure}[htbp]
\centering
\includegraphics[width=\linewidth]{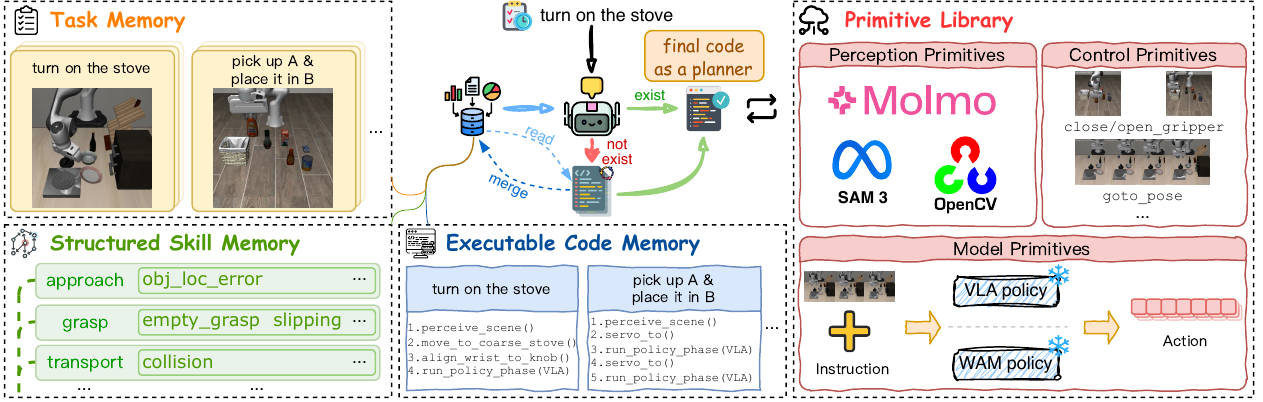}
\caption{The code framework. Knowledge is organized into three memories, a
\emph{Task Memory} of instructions, a \emph{Structured Skill Memory} of failure-to-repair skills, and an \emph{Executable Code Memory} of evolved code, around a central planner.
An instruction is matched to code that draws on a \emph{Primitive Library} of perception primitives (SAM3), control primitives (raw control APIs, e.g., grasp, motion), and model primitives (VLA, WAM, PPO), and that verifies each step before committing an action to the environment.}
\label{fig:code-framework}
\end{figure}

Fig.~\ref{fig:code-framework} shows this organization.
The framework stores what it has learned in three memories and what it can do in a primitive library, with code at the center tying the two together.
The \emph{Task Memory} holds the tasks the robot has been asked to carry out, such as ``turn on the stove'' or ``pick up A and place it in B.'' When a new task arrives, the agent searches the \emph{Task Memory}: on a hit it directly runs the code already stored in the \emph{Executable Code Memory}, and on a miss it draws on the \emph{Structured Skill Memory} and enters the closed-loop evolution described next (Section~\ref{sec:evolution}).

What that code invokes lives in the \emph{Primitive Library}, organized into three interchangeable kinds.
Perception primitives such as SAM3 ground the scene and answer \emph{what is where}.
Control primitives such as \texttt{goto\_pose} and \texttt{close/open\_gripper} supply the geometric motion, reaching a computed pose or closing a gripper, that a plan needs to execute.
Model primitives such as a generalist VLA ($\pi_{0.5}$)~\citep{pi05} or a predictive WAM (WorldDreamer)~\citep{worlddreamer} supply learned, contact-rich manipulation.
None of the three kinds is privileged over the others; the action model is one primitive among several, not the agent itself.

For an existing task, the code the coding agent has written serves as the planner.
It follows the plan step by step: it calls the perception primitives to ground the scene, reasons geometrically over what it perceives, decides from that which action to take, and only then invokes the action model to complete the task.
In this way the perception module compensates for the action model's weak perception, and code improves its planning ability.

Because the plan is written in this way, every decision the agent makes---which object, where, how, and whether the outcome was right---is a statement that can be read, tested, and edited, rather than a number hidden inside a policy's weights.
The central claim of this work follows: a physical AI is not an action model; it is perception, geometry, and action held together by an executable plan, of which the action model is only one, replaceable part.
This is the within-rollout half of the framework; the next subsection turns to the across-rollout half---how a plan is generated, diagnosed, and revised until it succeeds.

\subsection{Closed-Loop Evolution}
\label{sec:evolution}
\begin{figure}[htbp]
\centering
\includegraphics[width=\linewidth]{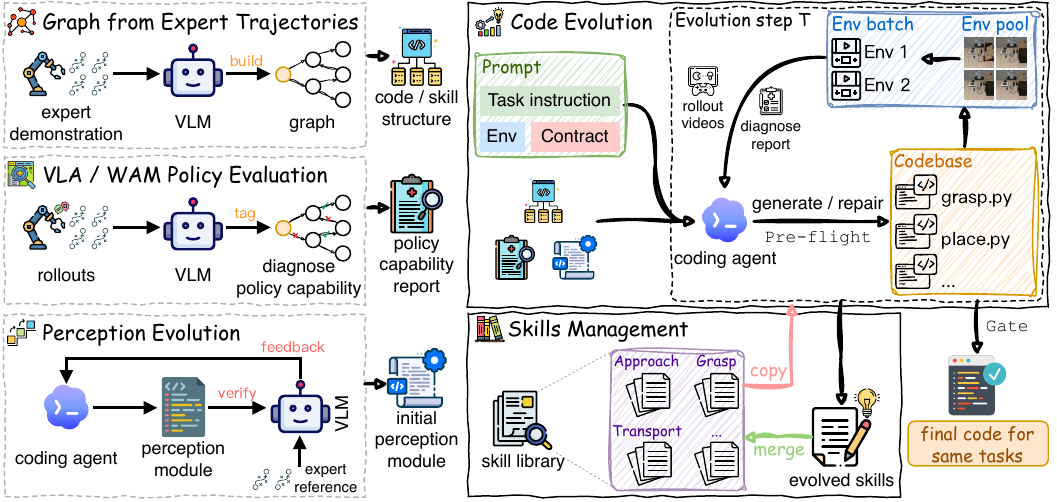}
\caption{The overall pipeline of \CodePhysical{}. (1) \emph{Optional}: a VLM reads
the expert-trajectory video and generates a workflow graph for the task, which fixes the structure of the code and the skills.
(2) \emph{Optional}: the VLA/WAM is rolled out on the task several times and a VLM judges at which workflow nodes the model errs and needs code takeover or repair, producing a capability assessment report.
(3) \emph{Optional}: for perception-dependent tasks, a coding agent calls SAM/Molmo to evolve perception code, verified against the expert trajectories by a VLM, yielding the initial perception module.
(4) The main evolution loop: given the task instruction, environment information, and the success contract, the coding agent builds the code package from the previously built code structure, runs it on multiple environments, and uses the diagnostics and rollout videos to decide the next edit; on errors it retrieves the matching skill from the skill library to repair the code, maintaining temporary in-progress skills, and on success it stores the evolved code in the code library and merges the evolved skills into the skill library.}
\label{fig:framework-arch}
\end{figure}

Section~\ref{sec:openloop} described the within-rollout half of the framework: how a plan, once written, executes open-loop against the scene.
It does not yet say where that plan comes from, or how it becomes correct.
That is the role of the closed loop.
\CodePhysical{} treats a task's code not as something a model writes once, but as a lineage that is generated, executed, diagnosed, and rewritten until it succeeds---and whose failures never leak into the code that already works.
Fig.~\ref{fig:framework-arch} shows the full pipeline.

\CodePhysical{} targets Physical AI embodiments that differ sharply from one another, from desktop robotic arms and household robots to robot vacuums, and each needs a different operating environment and simulation setup.
Some provide an expert-trajectory video, some must first have their action model's strengths assessed, and some lean heavily on perception.
Because no single pipeline accommodates all of these at once, the framework divides its evolution into \emph{required} and \emph{optional} steps: it keeps one core loop that every embodiment must pass through, and turns the embodiment-specific parts into steps that attach only when needed.

Three optional steps exist, each matching one kind of on-demand capability.
The first is workflow-graph generation: a VLM reads the expert-trajectory video and produces the task's workflow graph, which fixes the structure of the code and the skills.
The second is capability assessment: the VLA or WAM is rolled out on the task several times, and a VLM judges at which workflow nodes the model errs and needs code takeover or repair, producing a capability-assessment report.
The third is perception-module initialization: for perception-dependent tasks, a coding agent calls SAM or Molmo to evolve perception code, which a VLM verifies against the expert trajectories, yielding the initial perception module.
None of the three is mandatory; each turns on only when its input is present, namely expert trajectories, a VLA or WAM, or a perception-heavy task.

Perception code is evolved with the expert video doing more than verification: the VLM must also watch the trajectory to confirm which object each name refers to.
Consider the LIBERO-Object suite, where \emph{Alphabet soup} is a can-shaped object.
Without prior knowledge, even a strong VLM cannot recognize it reliably on every frame, but with the expert trajectory as reference the coding agent can write code that distinguishes \emph{Alphabet soup} from the other objects on the workbench, so recognition no longer depends on the VLM re-recognizing it correctly at each step.

There is a single required step, the main evolution loop, through which every embodiment and every task must pass.
Given the task instruction, environment information, and the success contract, the coding agent builds the code package on top of the previously built code structure, runs it across multiple environments, and uses the diagnostics and rollout videos to decide the next edit.
On errors it retrieves the matching skill from the skill library to repair the code; on success it stores the evolved code in the code library and merges the new skills into the skill library.
Whether or not the optional steps are enabled, every run converges onto this main loop.

\subsection{Evolution Mechanism}
\label{sec:evolution-mechanism}

A few terms are worth fixing up front.
Evolution usually targets a single task instantiated across many environment seeds: in reality the same task is executed repeatedly, but the environment shifts a little each time.
We evolve programs on development seeds and evaluate them using the benchmark-specific splits described in Section~\ref{sec:experiment-setup}.

Algorithm~\ref{alg:evolution} spells out the loop.
Two design choices are worth flagging.
First, the \emph{pre-flight} and \emph{probe} stages form a fast feedback loop that catches regressions without paying for a full batch.
Pre-flight runs as soon as the coding agent finishes writing code: it checks that the code runs at all, so a crash from an environment or driver interface is caught before it can be mistaken for a policy failure, and only code that passes pre-flight is evaluated in the environment.
Probe is a targeted repair check: the coding agent selects the three seeds that failed the previous round, fixes them, and re-verifies the fix on those three seeds alone; once they pass, the full training set is rolled out again to confirm the repair has not broken the seeds that were already correct.
Second, the self-review and the synthesis are a single call to the same writer, so the model that saw the evidence is the model that decides the next move.

\begin{algorithm}[htbp]
\caption{Closed-loop evolution of a task.}
\label{alg:evolution}
\KwIn{task description, environment information, success contract $C$, round budget $T$}
\KwOut{promoted code package $P$, distilled skills $S$}
Define success contract $C$ (scope, goal, terminal state, non-success signals)\;
$P \leftarrow$ build the code skeleton (from the task graph if available, else a single file)\;
$I \leftarrow$ task description\;
\For{$r \gets 1$ \KwTo $T$}{
  $P \leftarrow$ \textsc{Write}(Codex edits the candidate copy using $I$)\;
  \If{pre-flight or probe fails}{{fix the bug}\;}
  execute $P$ open-loop on every episode $\times$ every suite\;
  $(d, v) \leftarrow$ \textsc{Diagnose}(VLM node-wise verdict and video reward)\;
  $g \leftarrow$ \textsc{Gate}(deterministic milestone-prefix reward)\;
  \eIf{all rollouts succeed under $C$}{
    promote $P$ into the code library\;
    $S \leftarrow$ distill failure-to-repair pairs into the skill library\;
    \Return{$P$, $S$}\;
  }{
    $I \leftarrow$ \textsc{Synthesize}(Codex self-review of $d$ and $g$)\;
  }
}
\end{algorithm}

During evolution, defining what success is and how it is judged is critical.
Before evolution starts, on the first round, the framework writes a \emph{success contract}---the full evaluation scope (every suite, every episode), the authoritative goal condition, the graph's terminal state, and an explicit list of signals that do \emph{not} by themselves constitute success: a clean exit, a reached terminal state, a high dense reward, or the writer's own claim of convergence.
The same contract is injected into the writer, the diagnoser, and the reviewer on every round, so that what counts as success is never left to the model that is trying to succeed.

Beyond the final gating judgment and the phase-level checks inside the code itself, the loop also deploys a VLM (Gemini) to watch each task's rollout video and suggest plausible causes of its errors.
Two models split the loop (Table~\ref{tab:agents}) because the diagnosis signal must be independent of the model that writes: if a single model both wrote an edit and judged whether it worked, it would be grading its own exam.
The video diagnosis is a signal from a different modality and a different model, and the gate reward is a fallback that needs no model at all---a numeric signal that survives even if the video model misfires.

\begin{table}[htbp]
\centering
\setlength{\tabcolsep}{3pt}
\caption{Division of labor between the two models.}
\label{tab:agents}
\begin{tabular}{llll}
\toprule
Model & Role & Responsibility & Edits code \\
\midrule
GPT-5.6-sol & writer + synthesizer & edits phase code; self-review; next instruction & \ding{51} \\
Gemini-3.5-Flash & diagnoser & watches rollout video; node-by-node verdict vs.\ graph & \ding{55} \\
\bottomrule
\end{tabular}
\end{table}

After each round, the evolved code is rolled out across the entire training set to assess how it now performs.
The loop then reads the rollout output, both the code trace and the video feedback, and distills the cause of every remaining failure.
The video feedback concentrates on three questions: first, which failures were expected to have been fixed but recurred in this rollout; second, whether any regression appeared, where an environment seed that was previously correct now fails, and why; and third, what causes the errors that remain.
These answers are turned into the revision plan that drives the next round of evolution.

The product of a finished round is a code package.
Its perception code grounds the scene through SAM and Molmo; its planning code carries the geometric reasoning and motion control with NumPy, SciPy, Operational Space Control (OSC), and inverse-kinematics (IK) control; and for fine manipulation it delegates to the VLA.
Table~\ref{tab:hybrid} lays out this division: the deterministic phases run as fixed code, while the semantically hard phases such as grasp and place are handed to the VLA.

Evolution leaves behind more than the code package.
Throughout the run the framework records each problem it encountered and how it was fixed, and distills these into skills that are stored in the skill library for reuse across tasks.

\begin{table}[htbp]
\centering
\caption{The hybrid rollout. Deterministic phases run as fixed code; semantic
phases are delegated to the VLA.}
\label{tab:hybrid}
\begin{tabular}{ll}
\toprule
Phase & Executor \\
\midrule
grounding & fixed code (segmentation + geometry) \\
approach & fixed code (Cartesian servo, Pyroki IK rescue) \\
grasp & VLA \\
transport & fixed code \\
place & VLA \\
\bottomrule
\end{tabular}
\end{table}

What the loop leaves behind is the subject of the next subsection (Section~\ref{sec:skills}): the promoted code and the \emph{skills} distilled from its failures, and how the two are organized for reuse across tasks.

The writer, the pre-flight, and the rollouts all operate on an isolated candidate copy; the verified code library is never touched while a candidate is in flight.
Only after a full episode batch passes across every suite is the candidate promoted into the library atomically; on failure, or on exhaustion of the round budget, the library is left exactly as it was and the candidate remains in the run directory for diagnosis.
Candidate isolation prevents unverified edits from overwriting the stored program.

%% file: chapters/4_skill_library.tex
\subsection{Skill Library}
\label{sec:skills}

What the closed loop leaves behind is more than working code.
Each converged run also distills, from its failures, a record of what went wrong and what fixed it.
\CodePhysical{} accumulates these records into a \emph{skill library}.
A skill is a \emph{how-to-repair} note with two parts: the \emph{problem} it addresses---what failure to expect when attempting a given stage of a task---and the \emph{repair}---how to get past that failure once it happens.
The code library stores the executable plan; the skill library stores the experience of making that plan correct.

\begin{figure}[htbp]
\centering
\includegraphics[width=\linewidth]{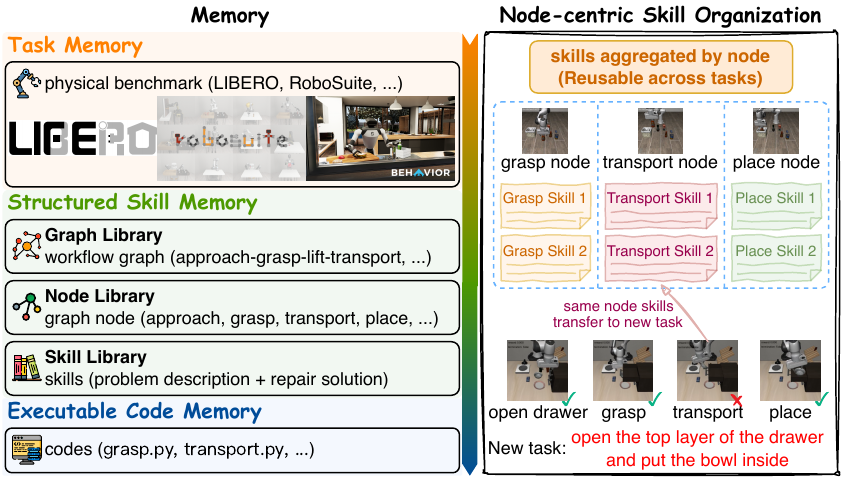}
\caption{How skills are stored and queried. (a) \emph{Storage}: at the top a
\emph{task memory} holds the descriptions of the tasks already evolved; under each task sits its \emph{structured skill memory}, and each task owns a workflow graph whose nodes each carry multiple skills describing how to repair the difficulties encountered at that node, with each skill pointing to the more detailed repair resources in the code memory.
(b) \emph{Query}: skills aggregate across tasks, so all skills under the same node are pooled together; when a new task hits a problem at an existing node (e.g., \texttt{transport}), it directly queries the skills stored under that node.}
\label{fig:task-library}
\end{figure}

\textbf{Storage.}
Skills are anchored on the task graph as a four-level structure (Fig.~\ref{fig:task-library}a): a \emph{graph library} (L1) holds the workflow graph of each task; each graph's nodes (L2)---grasp, transport, place, and so on---carry their own \emph{skill library} (L3), aggregating every problem--repair pair learned at that node; and the \emph{code library} (L4) sits alongside, holding the concrete repair code that each skill's \emph{repair} directly references.
A single node gathers several skills because the same stage recurs across many tasks and fails in many ways: a grasp node may accumulate one skill for a slipping grip, another for misalignment, another for an empty grasp.

A skill may include explanatory code fragments, while the maintained executable implementation resides in the code library.
Prior skill libraries often store each skill as an independent snippet of code, so that the library is itself a repository of executable procedures.
The repair note references the maintained implementation in the code library through the ``refers to'' edge in Fig.~\ref{fig:task-library}a. The skill stays a small, readable note, and the code remains in exactly one place, maintained by the same evolution loop that promoted it.
When a coding agent retrieves a skill, the reference leads it straight to the surrounding implementation it must edit, instead of asking it to regenerate that implementation from a copied fragment.

\textbf{Query.}
Retrieval is symmetric to storage: as Fig.~\ref{fig:task-library}b shows, a new task is not searched as a whole but first decomposed into the nodes its graph will traverse, and each node is then the key into the skill library.
In this sense the whole skill library is a \emph{node-centric} skill organization: skills are indexed not by task or by name but by the workflow node at which a given failure arises, so the node---not the task---is the unit that gathers and serves repairs.

These two choices---mounting skills on nodes and pointing them at shared code---buy the library its advantages.

\begin{figure}[htbp]
\centering
\includegraphics[width=\linewidth]{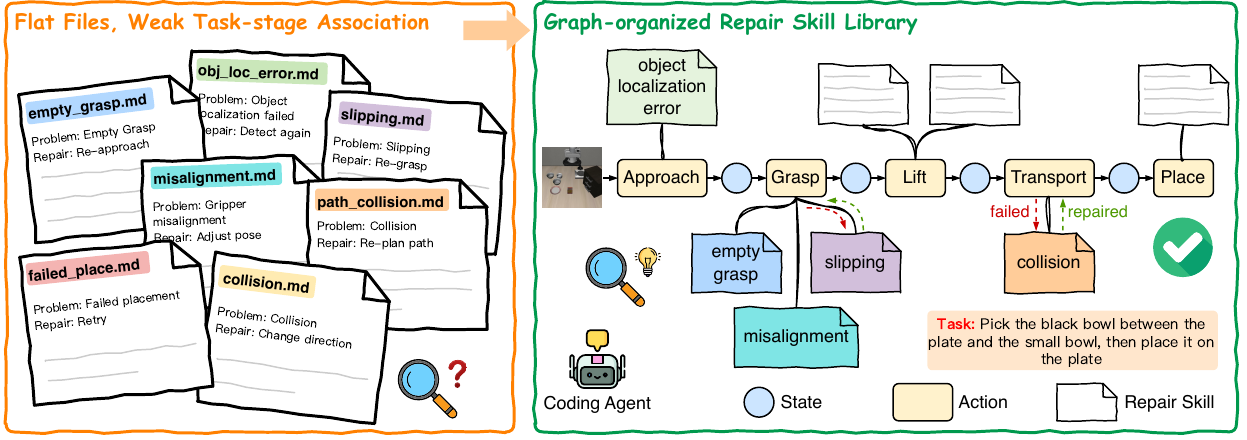}
\caption{Flat versus graph-organized skill storage. Left: writing each
``problem $\rightarrow$ repair'' pair to its own markdown file yields a heap of flat files with a weak association to the task stage.
Right: mounting each pair onto the graph node where it occurs (approach, grasp, transport, place) makes every skill retrievable by the stage of the task it belongs to.}
\label{fig:skill-graph}
\end{figure}

The first is \textbf{retrievability}.
The alternative, on the left of Fig.~\ref{fig:skill-graph}, is to write each problem--repair pair to its own markdown file.
That yields a heap of \texttt{*.md} files---\texttt{obj\_loc\_error.md}, \texttt{empty\_grasp.md}, \texttt{slipping.md}, \texttt{path\_collision.md}---with a weak association to the task stage at which each failure occurs, so a coding agent must search an unordered pile to find the one note relevant to the step it is debugging.
The right side shows the alternative \CodePhysical{} adopts: each problem--repair pair is mounted onto the node where it occurred---\texttt{approach}, \texttt{grasp}, \texttt{transport}, \texttt{place}.
An agent that has reached the \texttt{grasp} stage queries the \texttt{grasp} node and finds exactly the failures that happen during grasping and their repairs.
Node-based indexing groups repair notes by workflow stage to support targeted retrieval.

The second is \textbf{transfer}.
A novel task such as \emph{open the drawer} shares its grasp, transport, and place nodes with tasks the framework has already solved, so the skills mounted at those nodes transfer to it even though its own code has never been written.
This is the mechanism behind the how-to-repair transfer promised in Section~\ref{sec:why-code}: a repair learned while grasping a bowl is re-offered whenever a later task requires grasping, and Section~\ref{sec:experiment} shows the effect concretely---with the skill library, convergence arrives within the first few rounds.

Together with the code library, the skill library is what makes the closed loop cumulative.
The loop does not merely make one task correct; it converts the cost of each failure into a reusable, node-keyed repair, so that the next task---and every task after it---starts from what the framework has already learned.

%% file: chapters/4_model_feedback.tex
\subsection{Model Feedback}
\label{sec:model-feedback}

The loop generates, executes, diagnoses, and revises the code \emph{around} the VLA or WAM.
During the main harness-evolution loop, the underlying action model remains frozen.
That assumption buys a clean separation---the framework can be judged against a fixed, published checkpoint---but the loop remains limited by the capabilities and observability of its available perception, control, and learned-action primitives.
This subsection considers improving the action model through post-training.
Successful trajectories from a converged harness provide success-filtered demonstrations, which we use as expert data for fine-tuning on perturbed scenes.

The need is sharpest on benchmarks that offer no expert supervision.
LIBERO ships roughly fifty human-teleoperated demonstrations per task, the data a VLA is trained or fine-tuned on~\citep{libero}.
LIBERO-PRO is different: it perturbs those tasks---swapping object positions, changing the target---to test whether a model has merely memorized its training scenes, and it provides \emph{no demonstrations} for the perturbed variants~\citep{liberopro}.
There is, by construction, no expert trajectory to imitate for ``pick up the bowl after the plate and the ramekin have been swapped.''
These variants therefore lack benchmark-provided demonstrations for direct supervised adaptation; Section~\ref{sec:why-harness} illustrates failures under such perturbations.

\CodePhysical{} changes this picture because a converged code policy is itself a trajectory generator.
The loop of Section~\ref{sec:evolution} produces, for each perturbed task, a program that grounds, plans, and checks its way to success.
Run open-loop, that program emits precisely the data a policy learner needs: a sequence of observations and actions that reaches the goal in the perturbed scene, together with the hard success signal of the loop's gate.
The harness thus manufactures its own demonstrations where none exist---it needs no human to show how to re-grasp after a swap; the converged code already knows, and its rollouts record it.

Those rollouts are the expert data we feed back to the action model, and the route this work takes is \emph{supervised fine-tuning}: successful code rollouts are used directly as fine-tuning targets, distilling the code's explicit re-grounding and recovery behavior into the weights of the VLA or WAM.
The same trajectories could also drive \emph{reinforcement learning}, with the code policy's success gate as the reward and the model optimized to reach the states the code reaches.
Both routes aim to improve task performance under perturbations, but fine-tuning is the simpler, offline route, and the one we evaluate in Section~\ref{sec:harness-post-training}.
The experiments cover $\pi_{0.5}$ VLA fine-tuning across three LIBERO-PRO suites and DreamZero WAM fine-tuning on LIBERO-Object, evaluated under Swap and Task.

The benefit flows one way, and that is what distinguishes model feedback from a co-evolution of harness and model.
The harness is the data source, the model is the learner: the harness supplies demonstrations for perturbed scenes from its own successful rollouts, enabling model adaptation without new human labels for those trajectories.
Nothing flows back: the harness is not re-optimized against the improved checkpoint, and its code library, skill library, and converged programs are exactly what they were before fine-tuning.
Wrapping the harness back around the improved model and iterating is a natural extension, but not one this work undertakes.

%% file: chapters/5_experiment.tex
\section{Experiment}
\label{sec:experiment}

\subsection{Experimental Setup}
\label{sec:experiment-setup}

We evaluate \CodePhysical{} across seven benchmarks spanning four broad embodiment categories: robot arms (single-arm and bimanual manipulation), a humanoid upper body on a mobile base, a differential-drive vacuum, and a legged walking agent.
The settings also deliberately vary the action backend, a VLA ($\pi_{0.5}$), a WAM (WorldDreamer and DreamZero), a grasp detector (GraspNet), a pure control API, and a PPO policy, so that the experiments test the code framework itself rather than any single action model.
Throughout, the backends are frozen primitives and only the harness is evolved.
Evolution terminates when all development rollouts satisfy the success contract or the round budget is exhausted; only successful candidates are promoted, and crashes or timeouts count as failures.
Beyond the main results, which compare \CodePhysical{} against pure VLAs and WAMs, RL-based methods, and prior harness baselines, we further probe its zero-shot transfer to a new action model, its evolution dynamics, its runtime and monetary cost, and the evolved program serving as an expert-data collector whose successful trajectories are reflowed into post-training to improve the action model it wraps.

The evaluated subsets and backends are as follows:

\textbf{robosuite}~\citep{robosuite} (desktop robot arms, single-arm and bimanual).
We report seven tasks, namely 2A-Hand, 2A-Lift, Insert, Lift, Stack, Restack, and Wipe, with GraspNet~\citep{graspnet} providing the grasp action primitive.
We evolve one program per task on the held-out seeds 101-125 and report its success rate averaged over the 100 evaluation seeds 1-100.

\textbf{LIBERO}~\citep{libero} (desktop 7-DoF Franka Emika Panda arm).
We evaluate the Spatial, Object, and Goal suites, with ten tasks per suite, using the official LIBERO-finetuned $\pi_{0.5}$-LIBERO.
We evolve one program per task on seeds 0-14 and evaluate over 50 trials.

\textbf{LIBERO-PRO}~\citep{liberopro} (desktop robot arm under controlled perturbations).
We evaluate the same three suites under the Swap and Task perturbations, for 60 perturbed tasks in total, again with $\pi_{0.5}$-LIBERO (abbreviated as $\pi_{0.5}$).
Each perturbation is scored over 50 layout seeds and macro-averaged over the ten tasks in each suite; we evolve one program per task on seeds 0-14 and run that same program unchanged on all 50.
For LIBERO and LIBERO-PRO, the 50-seed evaluation includes the 15 seeds used for program evolution and 35 additional evaluation seeds, we follow the same setting in previous works.
The corresponding zero-shot transfer experiment replaces $\pi_{0.5}$-LIBERO with DreamZero, without re-evolving the harness.

\textbf{RoboCasa}~\citep{robocasa} (mobile manipulator).
We use WorldDreamer~\citep{worlddreamer365}, a video-conditioned WAM that maintains temporal world representations to drive a robot action policy. The main comparison is the Atomic-Seen setting reported as RoboCasa Target50 in the results table.
We report 20 seeds in a fixed-layout scene, with obstacles and distractors varying across seeds.

\textbf{BEHAVIOR-1K} (B1K)~\citep{b1k} (a humanoid household robot on a mobile base).
We use GraspNet for two household tasks, picking up a radio and picking up a soda can.
We evolve one program on seeds 26-35 and test it zero-shot on seeds 1-25, reporting navigation reachability separately from end-to-end task completion.
The adapted variant then re-evolves that program on the previously failing test seeds for eight rounds to repair manipulation errors.

\textbf{VacuSim}~\citep{vacusim} (differential-drive mobile robot vacuum).
The robot is controlled through a pure API of velocity commands and read-only sensor feedback, with cleaning progress over time as the outcome and roughly 80\% coverage as the attainable ceiling.

\textbf{MicroDuck}~\citep{microduck} (legged walking agent).
We use a PPO-trained Walker policy for the straight-line task.
The harness measures lateral drift and steers the same policy without retraining, and we report lateral error and endpoint distance rather than a binary success rate.

\subsection{Main Results}
\label{sec:experiment-results}

\subsubsection{robosuite}
\label{sec:results-robosuite}

\begin{figure}[htbp]
\centering
\includegraphics[width=\linewidth]{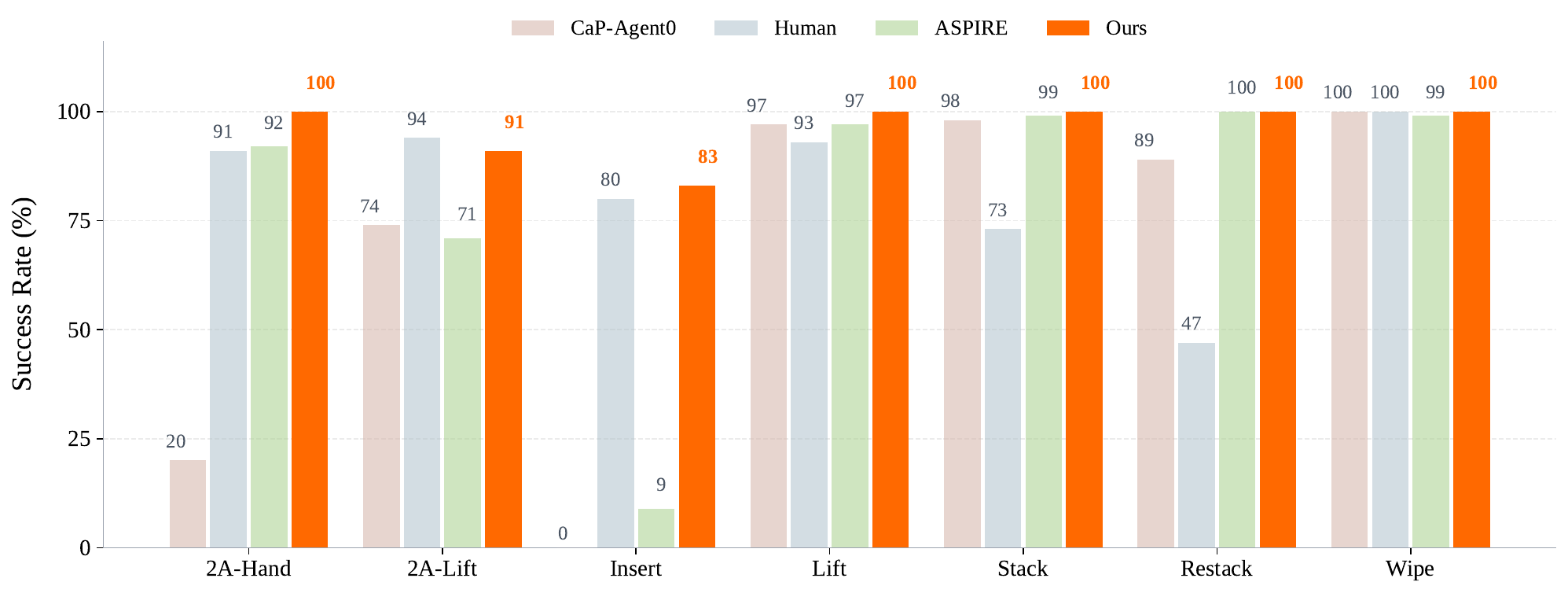}
\caption{Per-task success rate on robosuite. We report seven tasks: 2A-Hand, 2A-Lift, Insert, Lift, Stack, Restack, and Wipe, where 2A-Hand and 2A-Lift are bimanual (two-arm) tasks. For each task we evolve one program on the held-out seeds 101-125 and report its success rate averaged over the 100 evaluation seeds 1-100. CaP-Agent0, Human, and ASPIRE are taken from the ASPIRE paper~\citep{aspire}.}
\label{fig:robosuite-per-task}
\end{figure}

Fig.~\ref{fig:robosuite-per-task} reports per-task success on robosuite, with the underlying numbers tabulated in Table~\ref{tab:robosuite-per-task}.
\CodePhysical{} attains the highest average success rate of 96.3\%, clearing the strongest prior harness ASPIRE (81.0\%) by 15.3 points, the human baseline (82.6\%) by 13.7 points, and CaP-Agent0 (68.3\%) by 28.0 points.
The framework reaches a perfect 100\% on five of the seven tasks, namely 2A-Hand, Lift, Stack, Restack, and Wipe, and 91\% on the bimanual 2A-Lift, leaving Insert as its only task below 90\%.
The decisive gains concentrate on the two tasks that break fixed code scaffolds.
On Insert, a fine insertion that demands precise geometric reasoning, ASPIRE collapses to 9\% and CaP-Agent0 to 0\%, yet \CodePhysical{} holds 83\%, above even the human baseline of 80\%.
On 2A-Lift, \CodePhysical{} improves over ASPIRE by 20 points (91\% vs.\ 71\%).
Both tasks demand substantial perceptual alignment and geometric reasoning.
\CodePhysical{} organizes these functions into modular code packages to support inspection and revision.

\subsubsection{LIBERO}
\label{sec:results-libero}


\begin{table}[htbp]
\centering
\caption{Comparison of selected methods on the LIBERO benchmark (success rate, \%). We report success on the Spatial, Object, and Goal suites; Overall is the unweighted mean of the three displayed suite scores. Note that different papers adopt evaluation settings that are not entirely identical; our method uses the official $\pi_{0.5}$-LIBERO model (abbreviated as $\pi_{0.5}$), and the $\pi_{0.5}$ and DreamZero-LIBERO results are our own evaluation, with each task in every suite run over 50 seeds.}
\label{tab:libero}
\begin{tabular}{lcccc}
\toprule
Method & Spatial & Object & Goal & Overall \\
\midrule
\multicolumn{5}{l}{\textbf{Pure-Model}} \\
\midrule
DreamZero-LIBERO~\citep{dreamzerolibero} & 81.6 & 90.2 & 41.2 & 71.0 \\
TraceVLA~\citep{tracevla} & 84.6 & 85.2 & 75.1 & 81.6 \\
OpenVLA~\citep{openvla} & 84.7 & 88.4 & 79.2 & 84.1 \\
WorldVLA~\citep{worldvla} & 85.6 & 89.0 & 82.6 & 85.7 \\
CoT-VLA~\citep{cotvla} & 87.5 & 91.6 & 87.6 & 88.9 \\
MolmoAct~\citep{molmoact} & 87.0 & 95.4 & 87.6 & 90.0 \\
NORA~\citep{nora} & 92.2 & 95.4 & 89.4 & 92.3 \\
SmolVLA~\citep{smolvla} & 93.0 & 94.0 & 91.0 & 92.7 \\
GR00T N1~\citep{grootn1} & 94.4 & 97.6 & 93.0 & 95.0 \\
$\pi_{0.5}$~\citep{pi05} & 95.6 & \cellcolor{miBest}100.0 & 95.2 & 96.9 \\
$\pi_0$~\citep{pi0} & 96.8 & 98.8 & 95.8 & 97.1 \\
\midrule
\multicolumn{5}{l}{\textbf{RL-based}} \\
\midrule
World-Env~\citep{worldenv} & 87.6 & 86.6 & 86.4 & 86.9 \\
TGRPO~\citep{tgrpo} & 90.4 & 92.2 & 81.0 & 87.9 \\
GRAPE~\citep{grape} & 88.5 & 92.1 & 83.1 & 87.9 \\
VLA-RL~\citep{vlarl} & 90.2 & 91.8 & 82.2 & 88.1 \\
ThinkAct~\citep{thinkact} & 88.3 & 91.4 & 87.1 & 88.9 \\
$\pi_{0.5}$RLinf~\citep{pirl} & \cellcolor{miBest}99.0 & 96.0 & 97.0 & 97.3 \\
AtomVLA~\citep{atomvla} & 96.4 & \cellcolor{miSecond}99.6 & \cellcolor{miBest}97.6 & \cellcolor{miSecond}97.9 \\
\midrule
\multicolumn{5}{l}{\textbf{Harness}} \\
\midrule
OpenETA~\citep{openeta} (Code as Policy) & 8.0 & 26.0 & 21.0 & 18.3 \\
VLA-SCT~\citep{vlasct} (OpenVLA) & 91.2 & 92.8 & 82.0 & 88.7 \\
Harness VLA~\citep{harnessvla} ($\pi_{0.5}$) & 97.0 & \cellcolor{miBest}100.0 & 94.0 & 97.0 \\
\CodePhysical{} ($\pi_{0.5}$) & \cellcolor{miSecond}97.2 & \cellcolor{miBest}100.0 & \cellcolor{miSecond}97.2 & \cellcolor{miBest}98.1 \\
\midrule
$\Delta$ vs.\ $\pi_{0.5}$ & \textcolor{miBest}{+1.6} & \textcolor{miBest}{0.0} & \textcolor{miBest}{+2.0} & \textcolor{miBest}{+1.2} \\
\bottomrule
\end{tabular}
\end{table}

Table~\ref{tab:libero} reports per-suite and overall success rates on LIBERO, with methods grouped by family and ordered by increasing unweighted three-suite mean within each family.
We organize the comparison around three families of approaches.
\emph{Pure VLA} models, such as $\pi_0$~\citep{pi0}, OpenVLA~\citep{openvla}, SmolVLA~\citep{smolvla}, and GR00T N1~\citep{grootn1}, are end-to-end vision-language-action policies that map raw observations and language instructions directly to action tokens.
\emph{RL-based} models refine a VLA backbone with reinforcement learning or trajectory search, represented here by $\pi_{0.5}$RLinf~\citep{pirl} and AtomVLA~\citep{atomvla}.
\emph{Harness} methods wrap a policy, or replace it entirely, with an outer scaffold of code, memory, or self-correction.
Our harness baselines are VLA-SCT~\citep{vlasct} (over OpenVLA), Harness VLA~\citep{harnessvla} (a memory-guided agent that treats a frozen $\pi_{0.5}$ as a low-level primitive), OpenETA~\citep{openeta} (a pure code-as-policy method with no learned action model), and \CodePhysical{}, which, unlike the other three, retains a learned action model yet evolves and re-executes the surrounding code in a closed loop.

In Table~\ref{tab:libero}, most results are taken from the original papers, whereas DreamZero-LIBERO, $\pi_{0.5}$-LIBERO (abbreviated as $\pi_{0.5}$), and \CodePhysical{} are from our own evaluation. In the appendix~\ref{app:libero-details}, we provide more detailed per-suite and per-task results.

Among the three families, \CodePhysical{} achieves the best overall success rate reported in Table~\ref{tab:libero}: 98.1\%, which is 0.2 points above AtomVLA (97.9\%) and 1.1 points above Harness VLA (97.0\%).
It reaches a perfect 100.0\% on LIBERO-Object, tying $\pi_{0.5}$ and Harness VLA for the highest reported score, ranks second on LIBERO-Spatial at 97.2\% (within 1.8 points of $\pi_{0.5}$RLinf at 99.0\%), and ranks second on LIBERO-Goal at 97.2\% (within 0.4 points of AtomVLA at 97.6\%).
It is consequently the only method in the comparison to place in the top two across all three suites.

A further comparison is provided by OpenETA, a pure code-as-policy baseline without a learned action model.
OpenETA obtains 18.3\% overall success (8.0\%/26.0\%/21.0\% on Spatial/Object/Goal), whereas \CodePhysical{}, which combines evolved code with a learned $\pi_{0.5}$ action backend, obtains 98.1\%, a difference of 79.8 points.
This complete-system comparison is consistent with the value of the hybrid design, but it does not isolate the contribution of program evolution because the two systems use different action backends.

Both \CodePhysical{} and Harness VLA use a $\pi_{0.5}$ action primitive.
Under the evaluation settings reported in the table, \CodePhysical{} scores 1.1 points higher overall than Harness VLA (98.1\% vs.\ 97.0\%) and 3.2 points higher on LIBERO-Goal (97.2\% vs.\ 94.0\%).
Because the two results come from different evaluation protocols, this comparison should be interpreted as evidence of a performance advantage under the reported settings, rather than as a strictly controlled ablation.

\subsubsection{LIBERO-PRO}
\label{sec:results-libero-pro}

\begin{figure}[htbp]
\centering
\includegraphics[width=\linewidth]{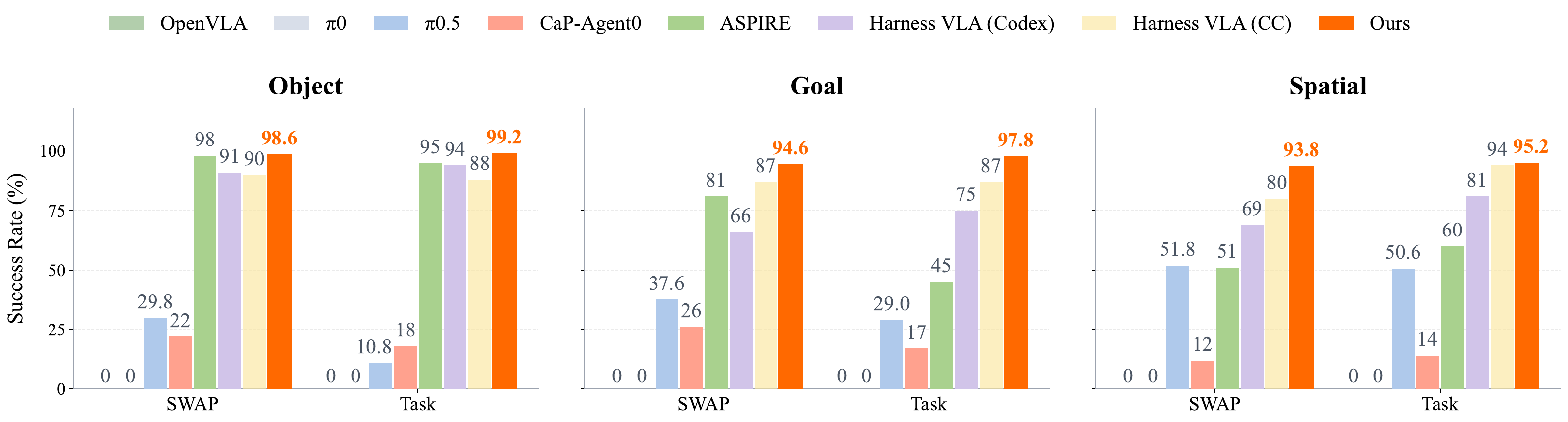}
\caption{Macro-averaged success rate reproduction on LIBERO-PRO. We evaluate the Object, Goal, and Spatial suites, each under the two perturbations Swap and Task, with 10 tasks per suite, 60 tasks in total. Success is macro-averaged: for each perturbation, the reported rate is the mean over its 10 tasks, each scored over 50 layout seeds (env seeds 0-49). For \CodePhysical{} we evolve one program per task on the first 15 of the 50 layout seeds (seeds 0-14), and then run that same program unchanged on all 50 seeds. The same program serves the base suite and both perturbations, so Swap and Task exercise it unchanged as the unperturbed task. OpenVLA, $\pi_0$, CaP-Agent0, and ASPIRE are taken from the ASPIRE paper~\citep{aspire}, where ASPIRE is likewise evaluated over 50 seeds per task; $\pi_{0.5}$ is evaluated by ourself; and the Harness VLA (Codex and CC) results are taken from the Harness VLA paper~\citep{harnessvla}, where each result is evaluated over 10 seeds per task.}
\label{fig:libero-reproduction}
\end{figure}

Standard LIBERO success can be misleading, however: a model can reach a high score by memorizing fixed action sequences and scene layouts rather than by understanding the task.
LIBERO-PRO~\citep{liberopro} probes exactly this.
It applies controlled perturbations to the LIBERO tasks, and we evaluate two of them, Swap and Task, to test whether a policy still succeeds when the memorized configuration no longer holds.
Swap is a position perturbation that exchanges the initial positions of two objects, forcing the model to re-anchor its actions to a changed layout; Task is a task perturbation that changes the task instruction given to the VLA model to test its ability to understand language.

The difficulty of these shifts is stark.
Models that exceed 90\% on standard LIBERO collapse under perturbation: in the original benchmark, OpenVLA and $\pi_0$ drop to 0.0\% on both Swap and Task~\citep{liberopro}, and even the strong $\pi_{0.5}$, which reaches 96.9\% on standard LIBERO, falls to 39.7\% on Swap and 30.1\% on Task in our evaluation.
Robustness here cannot be bought with a stronger backbone alone, which has motivated a family of harness-based remedies.

Two recent methods have attacked this gap with harness scaffolds.
ASPIRE~\citep{aspire}, a code-as-policy agent that writes, executes, and self-corrects robot programs against an explicit skill library, lifts LIBERO-PRO success to 72\% overall (77\% on Swap and 67\% on Task).
Harness VLA~\citep{harnessvla}, which treats a frozen VLA as a primitive and drives it with a memory-guided agent, reaches 87.7\% with its CC variant and 79.3\% with its Codex variant.
\CodePhysical{} extends this line of work and attains the best result, \textbf{96.5\%} overall (95.7\% on Swap and 97.4\% on Task), surpassing ASPIRE and the stronger Harness VLA CC variant by 24.8 and 8.8 points, respectively: because its closed-loop evolution corrects the code that frames each task, it recovers from the perturbations that break fixed scaffolds as thoroughly as it does on standard LIBERO.

\subsubsection{RoboCasa}
\label{sec:results-robocasa}

\begin{table}[htbp]
\centering
\caption{Results on the Atomic-Seen subset of RoboCasa Target50 (18 tasks; success rate, \%). Type distinguishes pure VLAs, World Action Models (WAM), and Harness methods. Note that different RoboCasa papers adopt different evaluation settings: for instance, Harness VLA runs 10 seeds per task on Atomic-Seen, whereas our method runs 20 seeds in a fixed-layout scene, with obstacles and distractors varying across seeds; to ensure a fair comparison, we evaluate both WorldDreamer and our method under a unified experimental setting.}
\label{tab:robocasa}
\begin{tabular}{lcc}
\toprule
Method & Type & Atomic-Seen \\
\midrule
$\pi_0$~\citep{pi0} & VLA & 34.6 \\
$\pi_{0.5}$~\citep{pi05} & VLA & 39.6 \\
RLDX-1~\citep{rldx1} & VLA & 60.0 \\
WorldDreamer~\citep{worlddreamer365} & WAM & 65.0 \\
Harness VLA (Codex)~\citep{harnessvla} & Harness & \cellcolor{miSecond}91.6 \\
Harness VLA (CC)~\citep{harnessvla} & Harness & 79.4 \\
\CodePhysical{}(WorldDreamer) & Harness & \cellcolor{miBest}92.2 \\
\midrule
$\Delta$ vs.\ WorldDreamer & & \textcolor{miBest}{+27.2} \\
\bottomrule
\end{tabular}
\end{table}

Table~\ref{tab:robocasa} reports results on RoboCasa, a distinct embodiment, the Franka Panda Emika robot, equipped with an Omron mobile base~\citep{haviland2022holistic}, paired with a different underlying action model, WorldDreamer.
\CodePhysical{} achieves 92.2\% on Atomic-Seen, compared with the reported 91.6\% for Harness VLA (Codex) under a different evaluation protocol.
These results are most informative when read against \CodePhysical{}'s own base policy: wrapping WorldDreamer with the closed-loop code framework lifts Atomic-Seen from 65.0\% to 92.2\% (+27.2 points).

Across the evaluated arm and mobile-manipulation settings, \CodePhysical{} improves task execution with both $\pi_{0.5}$ and WorldDreamer.
These results support the portability of the harness design across the tested action backends and control interfaces.

\subsubsection{BEHAVIOR-1K}
\label{sec:results-b1k}

\begin{figure}[htbp]
\centering
\includegraphics[width=\linewidth]{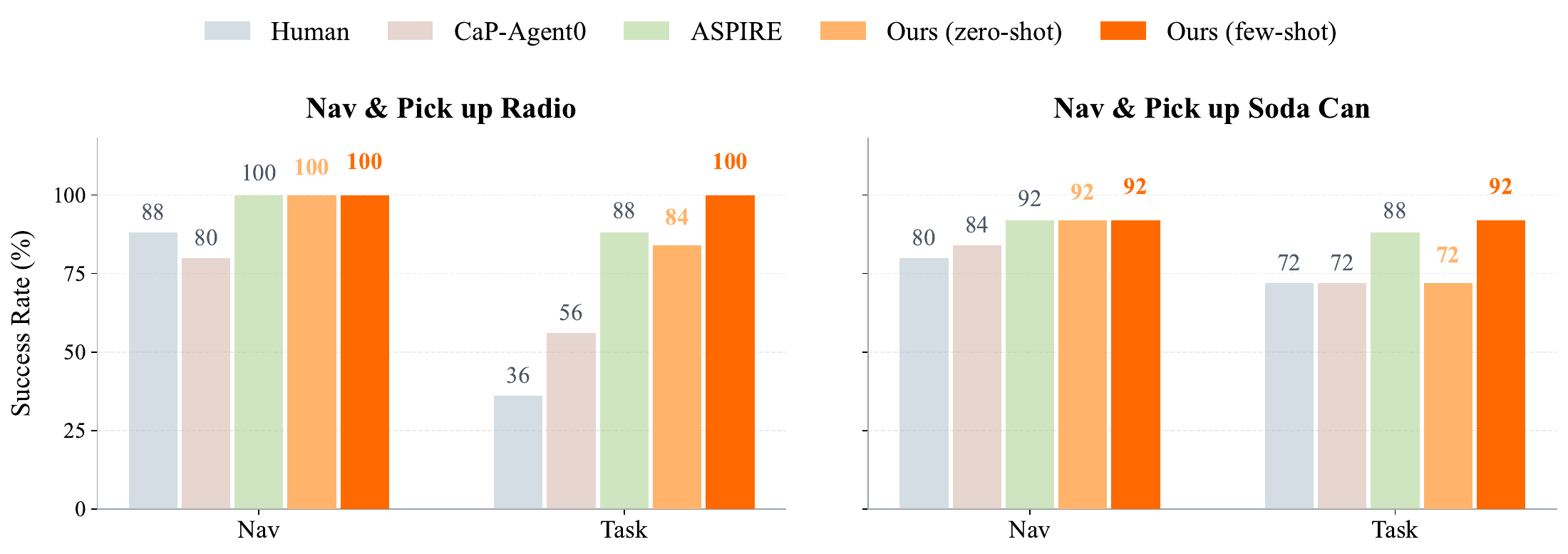}
\caption{A long-horizon humanoid household robot on a mobile base on two BEHAVIOR-1K household tasks. ASPIRE evolves one program per seed on seeds 26-35, distills ``how-to-repair'' skills from those runs, and then uses the skills to evolve one program per seed on seeds 1-25. \CodePhysical{} evolves one program from scratch on seeds 26-35 and distills ``how-to-repair'' skills; \emph{zero-shot} directly tests that program on seeds 1-25, while \emph{few-shot} re-evolves the zero-shot program on the failing seeds, where few-shot denotes eight iterations. Human, CaP-Agent0, and ASPIRE are taken from the ASPIRE paper~\citep{aspire}.}
\label{fig:b1k-mobile-manipulation}
\end{figure}

Fig.~\ref{fig:b1k-mobile-manipulation} reports results on two BEHAVIOR-1K household tasks using a humanoid upper body on a mobile base, separating navigation from end-to-end completion.
On picking up the radio, \CodePhysical{}'s \emph{zero-shot} program achieves 100\% navigation success, matching ASPIRE and beating CaP-Agent0 (80\%) and the human baseline (88\%), and its end-to-end task success of 84\% clears CaP-Agent0 (56\%) and human performance (36\%) while closing to within four points of ASPIRE (88\%).
The adapted variant then re-evolves the zero-shot program on the failing seeds for eight rounds and reaches 100\%, so that every navigable episode in the evaluated set is completed.
The soda-can task follows the same arc: navigation success is 92\% (above the human baseline of 80\% and CaP-Agent0 at 84\%), zero-shot task success ties CaP-Agent0 at 72\%, and adaptation lifts it to 92\%, four points above ASPIRE (88\%).
Across both tasks the navigation-to-completion gap, the episodes the base can reach but not finish, is eliminated entirely by the closed loop, which repairs the manipulation errors on whatever the mobile base can actually approach.

\subsubsection{VacuSim}
\label{sec:results-vacusim}

VacuSim is a model-free test of \CodePhysical{}: a differential-drive vacuum robot uses bumper and distance sensors together with a ROS~2 control API to maximize painted floor coverage within a time budget.
We evaluate a furnished $25\times25$ apartment, where recovery around furniture is the main challenge, and an unfurnished $40\times40$ apartment, where efficient long-range coverage dominates.
The score is computed by the ground supervisor rather than by planner-side tile counts.

\begin{figure}[htbp]
\centering
\includegraphics[width=\linewidth]{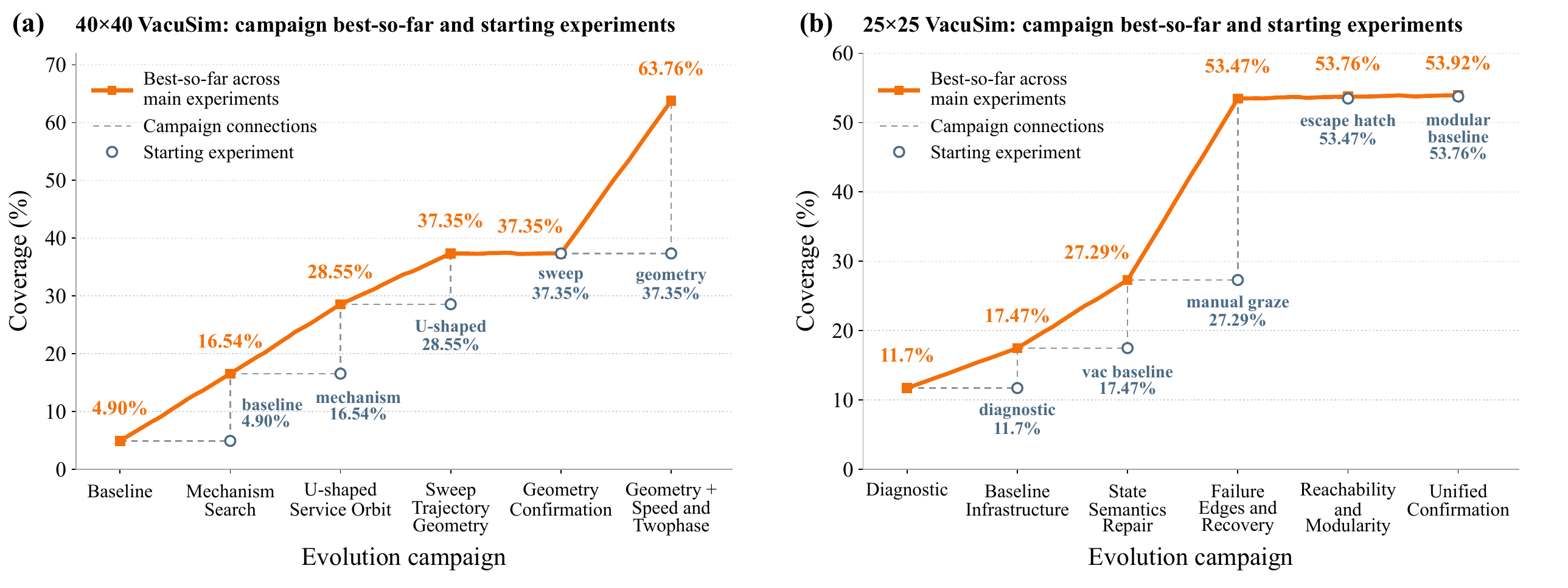}
\caption{Campaign-level evolution on VacuSim. Coverage improves from 4.90\% to 63.76\% on the unfurnished $40\times40$ map and from 11.7\% to 53.92\% on the furnished $25\times25$ map.}
\label{fig:vacusim-evolution}
\end{figure}

Across the campaign summarized in Fig.~\ref{fig:vacusim-evolution}, the evolution is hierarchical: each outer round selects one technical direction, which is explored through smaller measurable directions from isolated code candidates.
Every direction maintains its own Skill and Code memory; validated improvements are retained with their diagnostics, while failed candidates remain as negative evidence without overwriting the champion.
After three valid attempts without improvement, a direction is retired and validated skills are tested in bounded combinations.
If these combinations also plateau, the next outer round moves to another direction while preserving the previous libraries for retrieval.

\begin{figure}[htbp]
\centering
\includegraphics[width=\linewidth]{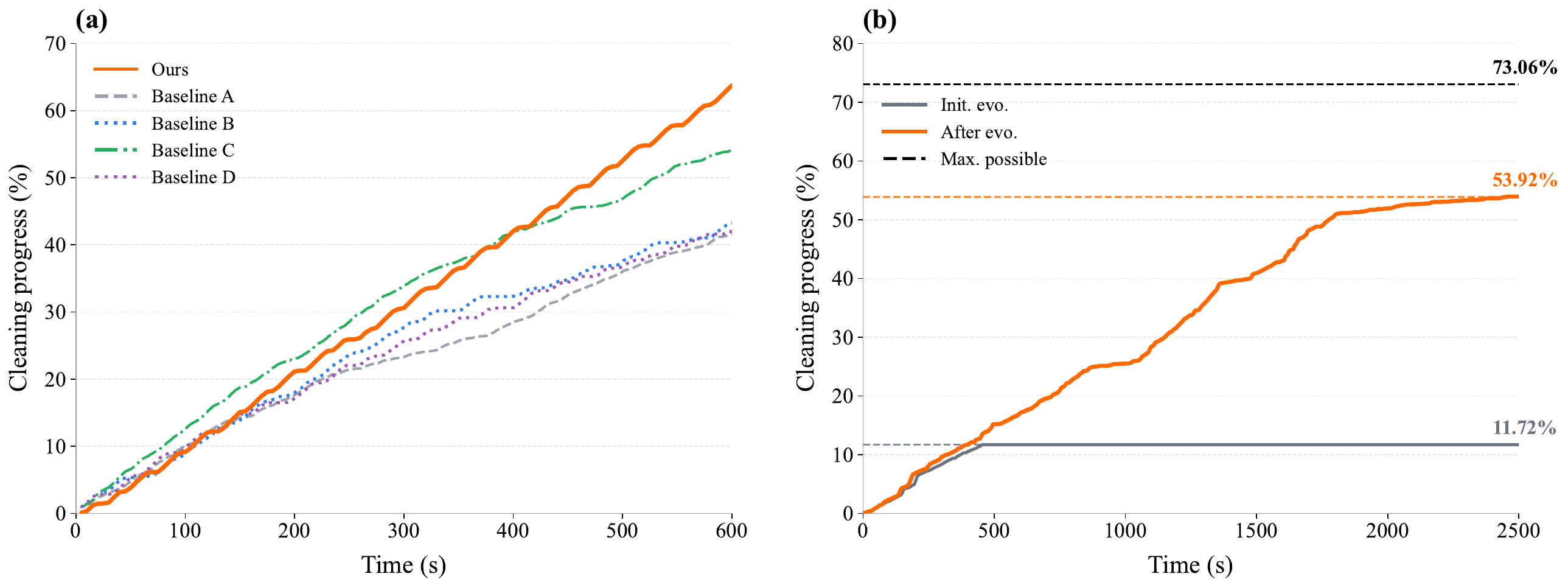}
\caption{Cleaning progress over time on VacuSim. Panel (a) compares the final \CodePhysical{} policy with four baselines~\citep{mints2023ros2webots}
on the unfurnished $40\times40$ apartment, while panel (b) compares the initial and evolved policies on the furnished $25\times25$ apartment. The final \CodePhysical{} policy reaches 63.76\% coverage on the $40\times40$ map and improves from 11.72\% to 53.92\% on the $25\times25$ map; the 73.06\% dashed line denotes the maximum physically cleanable coverage of the furnished map.}
\label{fig:vacusim}
\end{figure}

Fig.~\ref{fig:vacusim} reports the resulting cleaning progress over time, comparing the final policy against four baselines on the unfurnished $40\times40$ map and the initial against the evolved policy on the furnished $25\times25$ map.
VacuSim therefore instantiates the main evolution loop without the optional VLA, workflow-graph, or perception stages: diagnostics and painted coverage guide code revision, while validated direction libraries provide reusable repair knowledge.
The final policy executes without an LLM call on the critical path.

\subsubsection{MicroDuck}
\label{sec:results-microduck}

\begin{figure}[htbp]
\centering
\includegraphics[width=\linewidth]{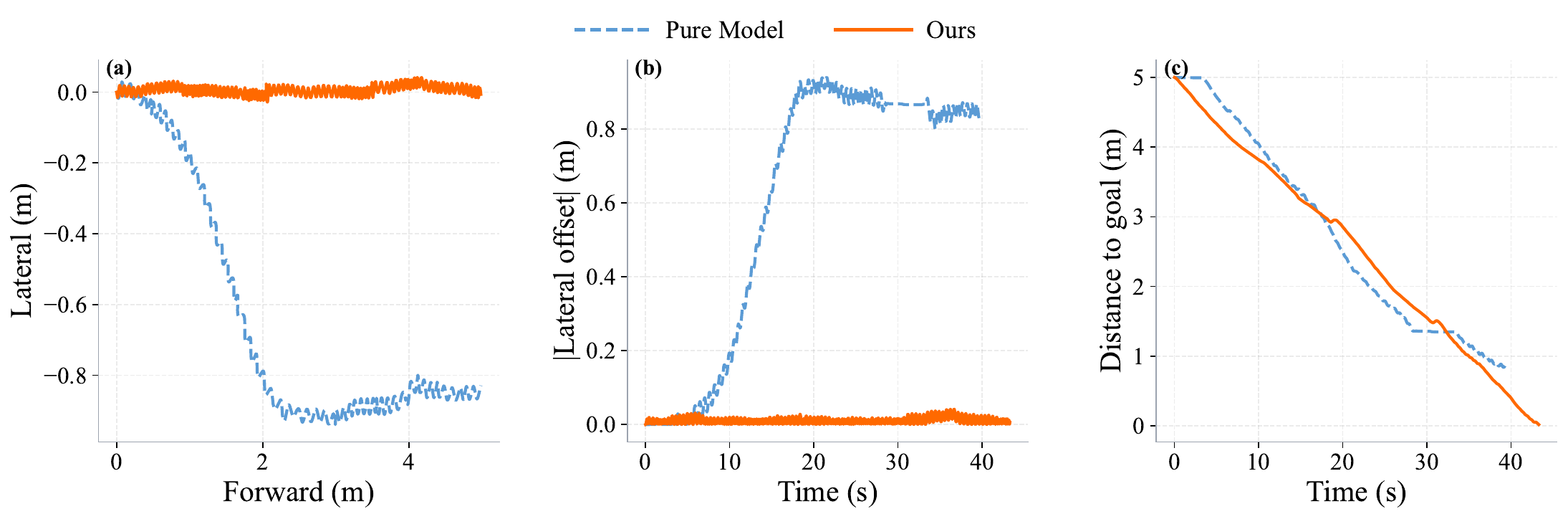}
\caption{Performance of the pure PPO model and the same model with the \CodePhysical{} harness on MicroDuck's straight-line walking task: (a) the planar trajectory, showing how the lateral error evolves as forward progress increases; (b) the absolute lateral error over time; and (c) the distance between MicroDuck and the goal over time.}
\label{fig:microduck-comparison}
\end{figure}

MicroDuck is a walking agent whose objective is simply to walk forward in a straight line.
The task is more demanding than it appears: a PPO-trained Walker model advances toward the goal, but nothing in its training objective penalizes sideways drift, so it veers off course as it walks.
Fig.~\ref{fig:microduck-comparison} contrasts the pure PPO model against the same model wrapped by the \CodePhysical{} harness across three views of the rollout.

Panel (a) plots the planar trajectory: over 4~m of forward progress the pure model drifts up to $0.8$~m off the ideal line, whereas the harness holds its lateral error near zero.
Panel (b) shows the absolute lateral error over time, which accumulates for the pure model but remains flat for the harness.
Panel (c) tracks the distance to the goal, which the pure model, carried off course by the accumulated drift, fails to close, while the harness drives it to zero and reaches the endpoint.
The correction comes from code rather than retraining: the harness measures the heading error and steers the same underlying Walker policy back onto the line, so, as elsewhere in this work, the gain is obtained without retraining the Walker policy.

\subsection{Harness Zero-shot}
\label{sec:harness-zeroshot}

The results so far pair each benchmark with the action model against which its harness was evolved.
Here we ask a stricter question: does the harness transfer to a different action model without any retraining?
To test this, we take the harness evolved on LIBERO-Object with $\pi_{0.5}$ as the underlying policy and apply it \emph{zero-shot} to DreamZero, a World Action Model (WAM) that jointly predicts future scene evolution and action chunks rather than acting on the current observation alone.
Nothing in the harness is re-evolved or re-tuned for the new model; the only change is that every action-model call now invokes DreamZero instead of $\pi_{0.5}$.

\begin{figure}[htbp]
\centering
\includegraphics[width=\linewidth]{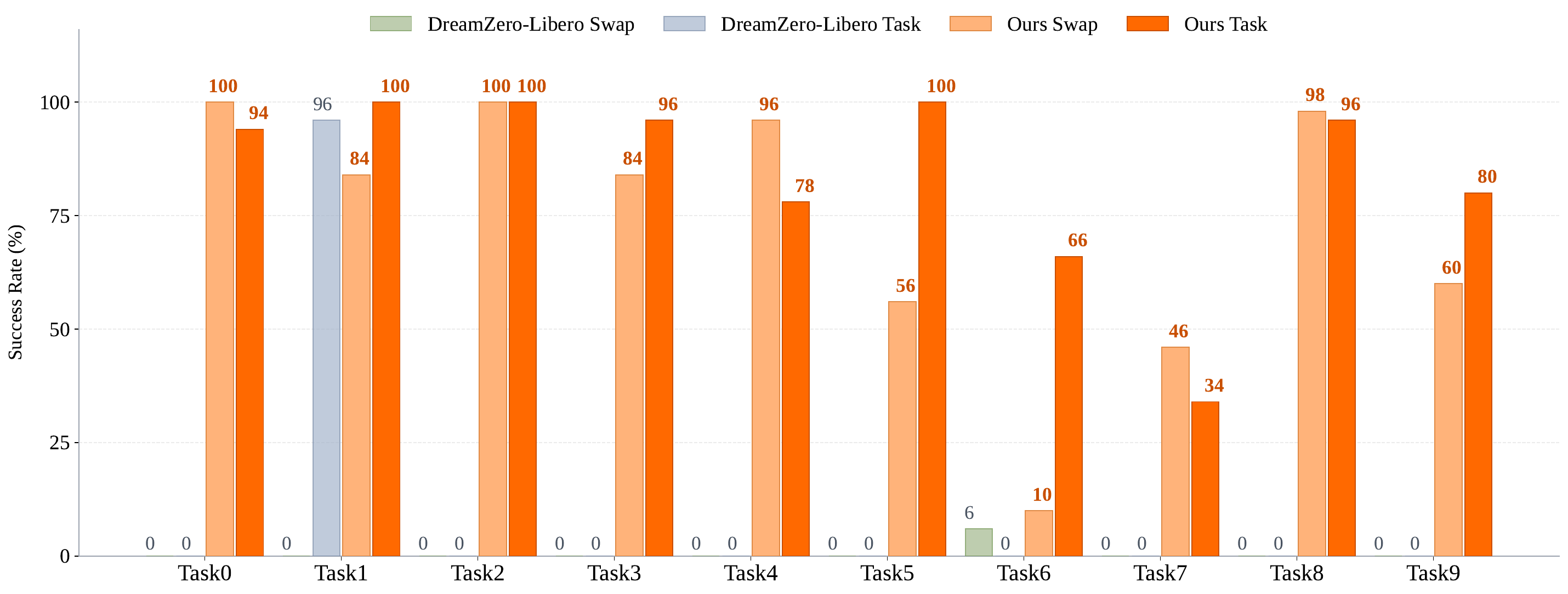}
\caption{Zero-shot transfer of the \CodePhysical{} harness to DreamZero, a World Action Model. Per-task success rate (\%) on the LIBERO-Object suite under the Swap and Task perturbations of LIBERO-PRO, comparing DreamZero-Libero against the \CodePhysical{} harness applied \emph{zero-shot}.}
\label{fig:harness-zeroshot}
\end{figure}

Fig.~\ref{fig:harness-zeroshot} reports per-task success under the Swap and Task perturbations of LIBERO-PRO~\citep{liberopro}.
The original DreamZero WAM barely generalizes: it scores 0.6\% overall on Swap, with its only successes being 3 of 50 on the butter task, and 9.6\% on Task, a total carried almost entirely by a single task (cream cheese$\to$alphabet soup, at 96\%), a memorization artifact in which the model recognizes the alphabet-soup object it has already seen and replays the corresponding behavior.
Applying the $\pi_{0.5}$-evolved harness \emph{zero-shot} lifts the same model to 73.4\% on Swap and 84.4\% on Task.
The gains come from the code around the action model rather than the action model itself: the harness supplies the perception, geometric reasoning, and verification that DreamZero's weights do not, and because those components are written as executable code rather than trained parameters, they carry over to a new action model untouched.
The residual gap relative to the 96.5\% that the same harness attains over $\pi_{0.5}$ on LIBERO-PRO reflects the weaker underlying WAM, but the direction of transfer is the point: code generalizes where weights do not.

\subsection{Evolution Process}
\label{sec:evolution-process}

\begin{figure}[htbp]
\centering
\includegraphics[width=\linewidth]{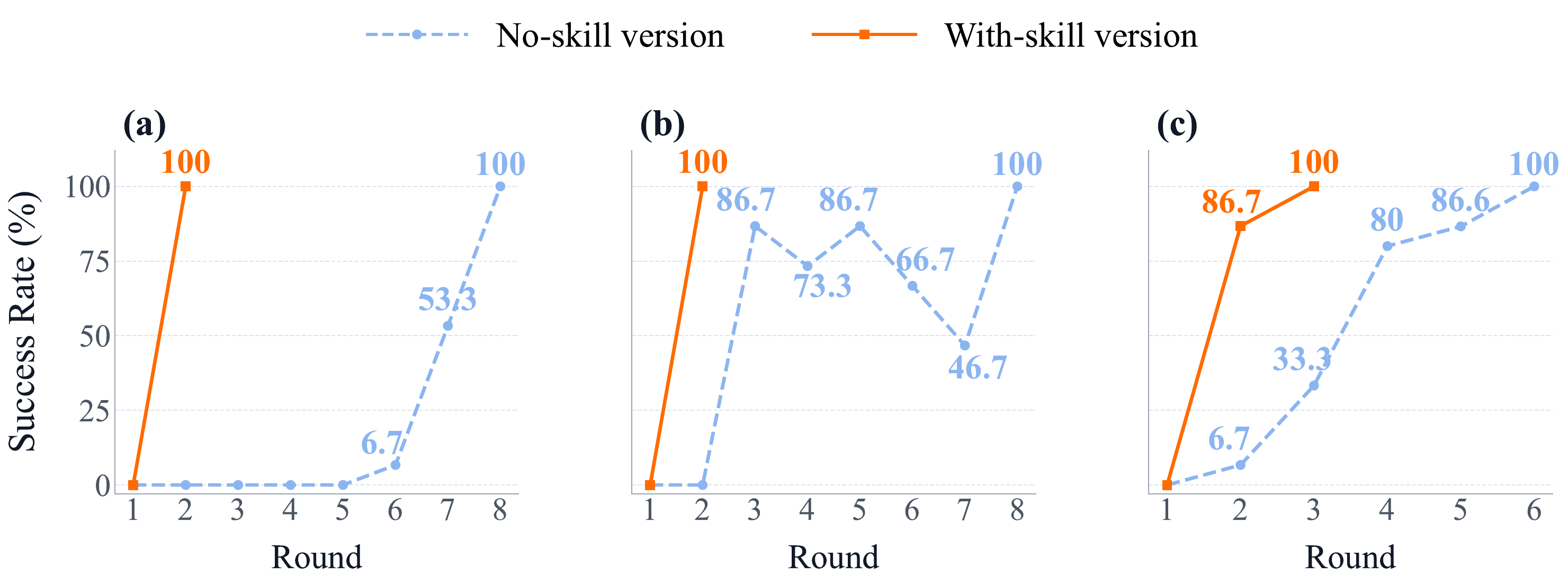}
\caption{The average results of closed-loop evolution of \CodePhysical{} on LIBERO Object, LIBERO-PRO Object Task and Swap. We evolve the task code under two conditions: with no access to the skill library (\emph{no-skill}) and with the evolved skill library (\emph{with-skill}), and plot success rate per round for three representative tasks: (a) Task~0, (b) Task~1, and (c) Task~5. Success rate is the number of successful rollouts summed over seeds 0-14 across the Swap and Task perturbations, 30 seeds in total.}
\label{fig:evolution-process}
\end{figure}

Fig.~\ref{fig:evolution-process} traces how the closed loop converges in practice.
On LIBERO-PRO we evolve one program per task over the 15 development scenes, and then run that same program unchanged against the base task, the Swap position perturbation, and the Task perturbation~\citep{liberopro}.
Evaluating each candidate across multiple development conditions checks whether an edit preserves task performance across the tested layouts and perturbations.
The figure isolates the LIBERO-Object suite and plots three representative tasks, namely task~0, task~1, and task~5, each with two curves: a \emph{no-skill} baseline, in which the code is evolved without access to the skill library, and a \emph{with-skill} variant that draws on it.

Without skills, the three tasks follow three qualitatively different trajectories.
Task~0 (Fig.~\ref{fig:evolution-process}a) exhibits an \emph{``Aha'' moment}: success stays pinned near zero for several rounds while no single edit resolves the failure, and then the round that removes the blocking error sends it to 53.3\% and on to 100\%.
Task~1 (Fig.~\ref{fig:evolution-process}b) meanders instead: its success oscillates between 86.7\%, 73.3\%, 86.7\%, 66.7\%, and 46.7\% before finally landing on the correct program, indicating a rugged search landscape in which the code wanders before converging.
Task~5 (Fig.~\ref{fig:evolution-process}c) climbs monotonically, accumulating partial corrections (6.7\%$\to$33.3\%$\to$80\%$\to$86.6\%) until it reaches 100\%.
With the skill library, by contrast, all three jump to 100\% within the first two rounds, which is the clearest evidence of the skills' value: they supply the correct manipulation primitives up front, so the closed loop spends its budget refining rather than re-discovering elementary behaviors.

Not every task exhibits an ``Aha'' moment.
The ``Aha'' moment indicates a bottleneck: once it is resolved, success rises rapidly.
Some tasks instead undergo a meandering exploration in the absence of skills before they eventually find the correct answer, and some tasks are limited by several interacting factors, so that reaching 100\% still requires two or three rounds even with the skill library.

\begin{figure}[htbp]
\centering
\includegraphics[width=\linewidth]{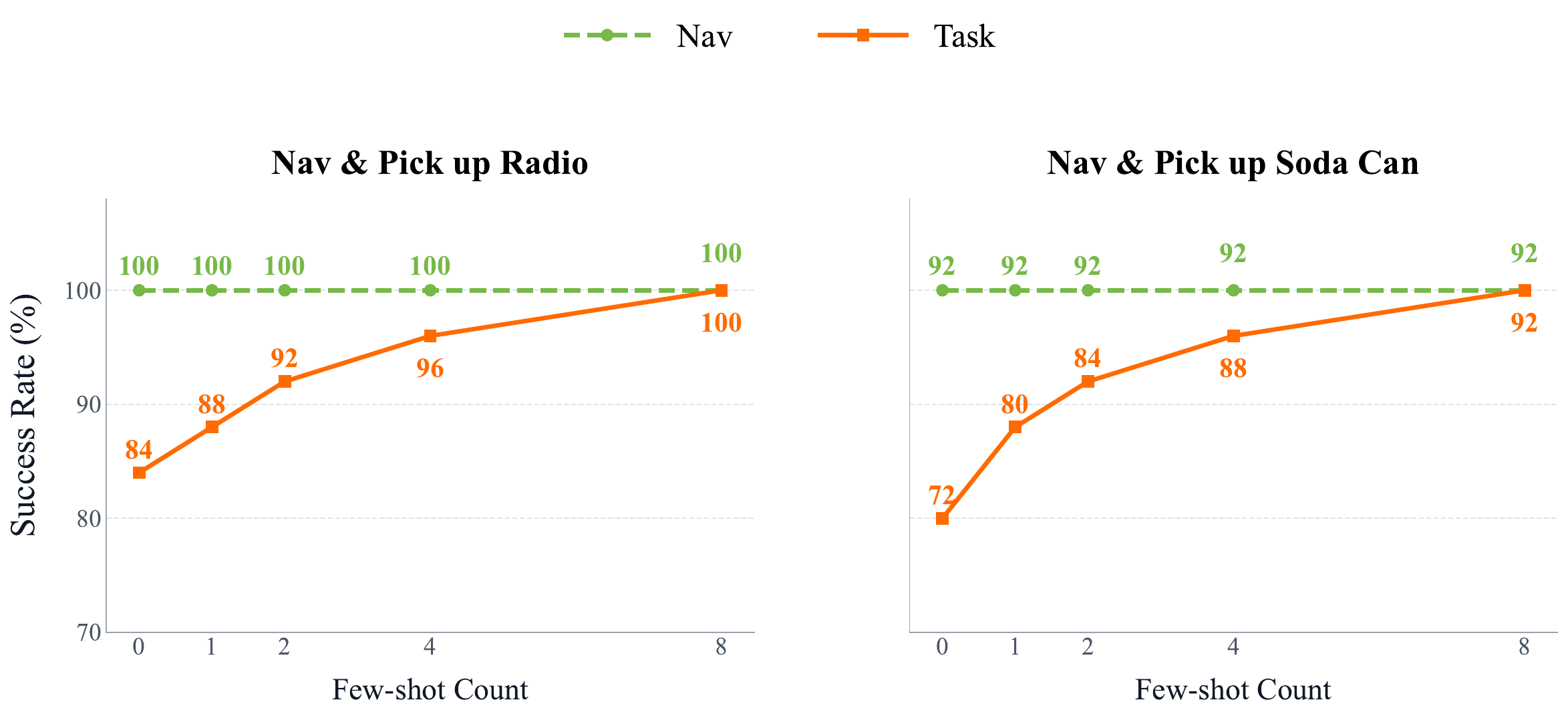}
\caption{Few-shot evolution on BEHAVIOR-1K. On the two household tasks, picking up a radio and picking up a soda can, \CodePhysical{} re-evolves the zero-shot program on the failing seeds 1-25, and we separate the navigation stage from the full task: \emph{Nav} measures how often the mobile base reaches the target object at all, while \emph{Task} measures end-to-end success. Navigation is flat (100\% and 92\%), marking the set of episodes the agent can actually reach, and the task success rate climbs with each few-shot round (84\%$\to$100\% and 72\%$\to$92\%) until it converges to that ceiling.}
\label{fig:nav-task-fewshot}
\end{figure}

Fig.~\ref{fig:nav-task-fewshot} shows the same closed loop on the mobile-manipulation benchmark.
On BEHAVIOR-1K we re-evolve the program on the seeds that failed the zero-shot pass (seeds 1-25), and here we isolate navigation from the full task to make the effect legible.
Navigation success remains at 100\% for the radio task and 92\% for the soda-can task throughout the reported adaptation rounds.
The \emph{Task} curve starts below the corresponding navigation success rate (84\% and 72\%) and then climbs monotonically with the number of adaptation rounds, reaching 100\% and 92\% after eight rounds.
In the reported final evaluation, every episode in which navigation succeeds also reaches task completion.

\subsection{Cost}
\label{sec:cost}

\begin{figure}[htbp]
\centering
\includegraphics[width=\linewidth]{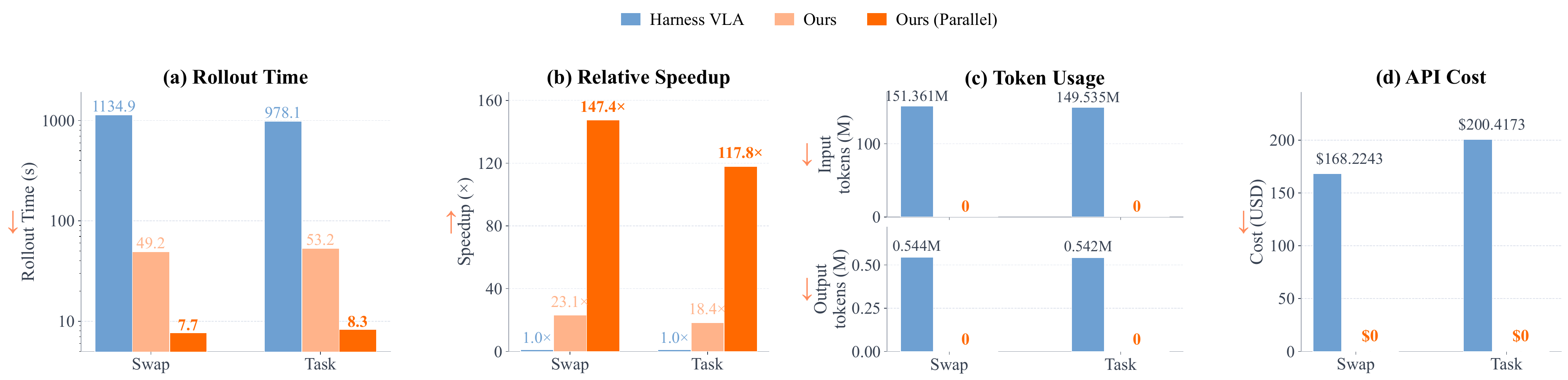}
\caption{Cost on LIBERO-Goal Task~1 under the Swap and Task perturbations, each evaluated over seeds 0-49 (100 seeds in total). (a) Rollout time. (b) Relative speedup. (c) Online high-level LLM token usage; Harness VLA uses GPT-5.6-sol with reasoning effort \emph{xhigh}. (d) Online high-level LLM API charges. Because Harness VLA queries GPT for an instruction at every action step, GPT's input (the image and current state) dominates the token count and its deliberation dominates the time. Since \CodePhysical{} runs many tasks in parallel across GPUs, its parallel variant is the result of 7 GPUs rolling out simultaneously.}
\label{fig:cost}
\end{figure}

We measure runtime cost in two regimes.
The serial timing is recorded over a complete run of 50 rollouts per task for both Harness VLA and \CodePhysical{}, and it covers the entire execution loop, including environment startup, perception, and control, rather than only the forward pass of the policy.
On top of this, \CodePhysical{} reports a parallel variant obtained through our own infrastructure, which runs perception, action, and environment simulation concurrently on 7 GPUs and overlaps their execution.
Fig.~\ref{fig:cost} therefore separates three configurations per task: Harness VLA (serial), \CodePhysical{} serial, and \CodePhysical{} parallel.

Fig.~\ref{fig:cost}a reports rollout time.
Harness VLA requires 1134.9~s on Swap and 978.1~s on Task  per rollout,  whereas \CodePhysical{} only need 49.2~s and 53.2~s per rollout on Swap and Task suite, and the parallel variant cuts this further to 7.7~s and 8.3~s.
Relative to serial Harness VLA, \CodePhysical{} achieves  speedups of 23.1$\times$ on Swap and 18.4$\times$ on Task in serial execution, and 147.4$\times$ and 117.8$\times$ in the separate seven-GPU configuration (Fig.~\ref{fig:cost}b).

The gap is architectural.
During execution Harness VLA must invoke an agent at every step, and in our reproduction that agent is Codex, a large reasoning model that reads the observation and memory and deliberates before issuing each action.
This per-step reasoning is an in-the-loop dependency: it serializes the rollout and adds the latency of a large model to every step.
\CodePhysical{}, in contrast, compiles a task into code once and then executes that code open-loop, without online high-level LLM deliberation on the critical path. The same dependency drives the token usage in Fig.~\ref{fig:cost}c. Across its 50 rollouts, Harness VLA consumes 151.36M input tokens and 0.544M output tokens on Swap, and 149.54M input tokens and 0.542M output tokens on Task. According to the official pricing of GPT-5.6, Harness VLA consumes a total of \$168.2243 on Swap (input + output + cache hits), and a total of \$200.4173 on Task. Since our code does not need to spend money calling a coding agent for reasoning during execution, both the tokens we consume and the amount we spend are 0.

%

Fixed-program \CodePhysical{} execution incurs no online high-level LLM API charge.
This does not imply zero cost for perception, action-model inference, simulation, or hardware, and the execution-stage comparison does not establish total program-development or lifecycle costs.

\subsection{Harness Trajectories for Post-Training}
\label{sec:harness-post-training}

The closed loop of Section~\ref{sec:evolution} produces successful trajectories under the perturbations that challenge the underlying action model.
A hard success gate retains only rollouts satisfying the environment's task-success condition, yielding success-filtered demonstrations for fine-tuning.
Once converged, the program collects trajectories without online high-level LLM calls.
This provides demonstrations for perturbed LIBERO variants, for which the benchmark supplies no human-teleoperated counterparts~\citep{liberopro}.
This provides demonstrations for perturbed LIBERO variants, on which VLA and WAM models often struggle~\citep{liberopro}.
We evaluate whether these demonstrations improve the underlying action model through two studies: VLA fine-tuning with $\pi_{0.5}$ and WAM fine-tuning with DreamZero.

\begin{figure}[t]
\centering
\includegraphics[width=\linewidth]{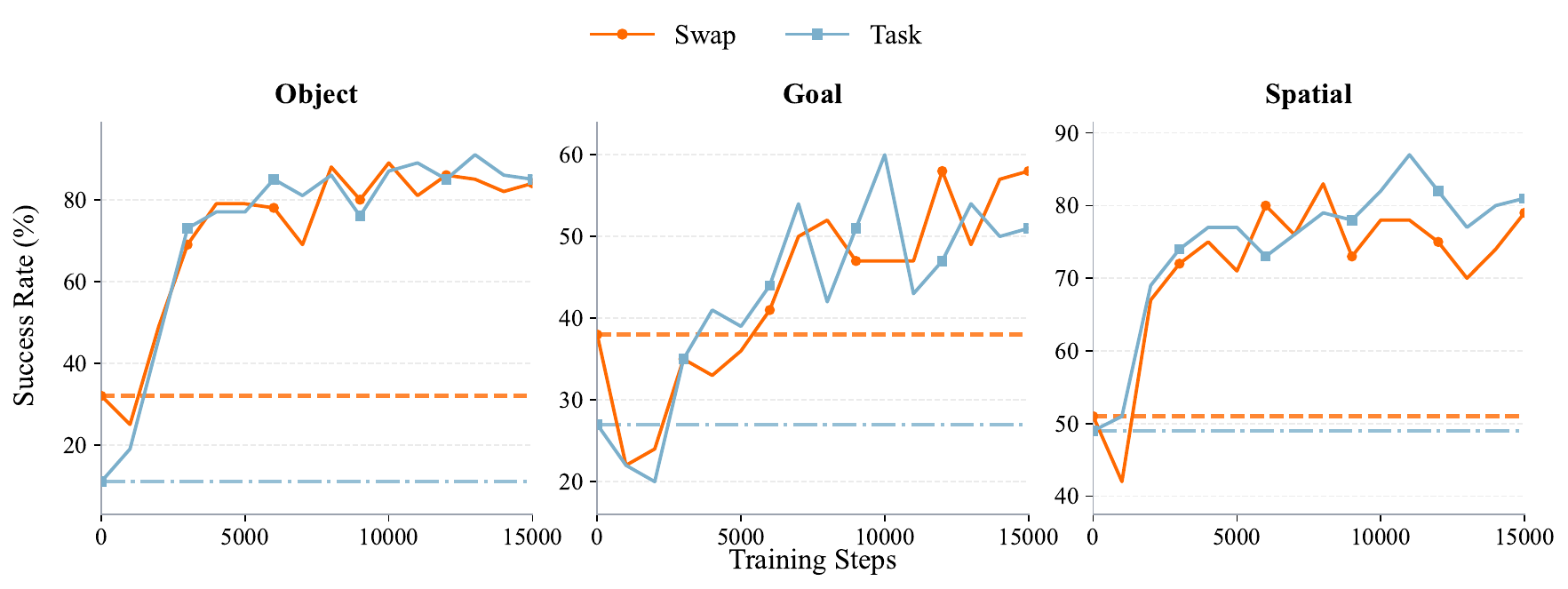}
\caption{Per-suite VLA fine-tuning on harness trajectories. Each $\pi_{0.5}$-LIBERO model is fine-tuned on one suite (Object, Goal, or Spatial), mixing Base, Swap, and Task trajectories equally (1:1:1). Solid curves show raw checkpoint test success rates (\%) against fine-tuning steps under the Swap and Task perturbations, with markers every third checkpoint. Dashed horizontal lines mark the original $\pi_{0.5}$-LIBERO policy's pre-fine-tuning success rate for each suite and perturbation. For each task, seeds 0-39 form the training set and seeds 40-49 the test set.}
\label{fig:harness-post-training-suites}
\end{figure}

\begin{figure}[!t]
\centering
\includegraphics[width=0.83\linewidth]{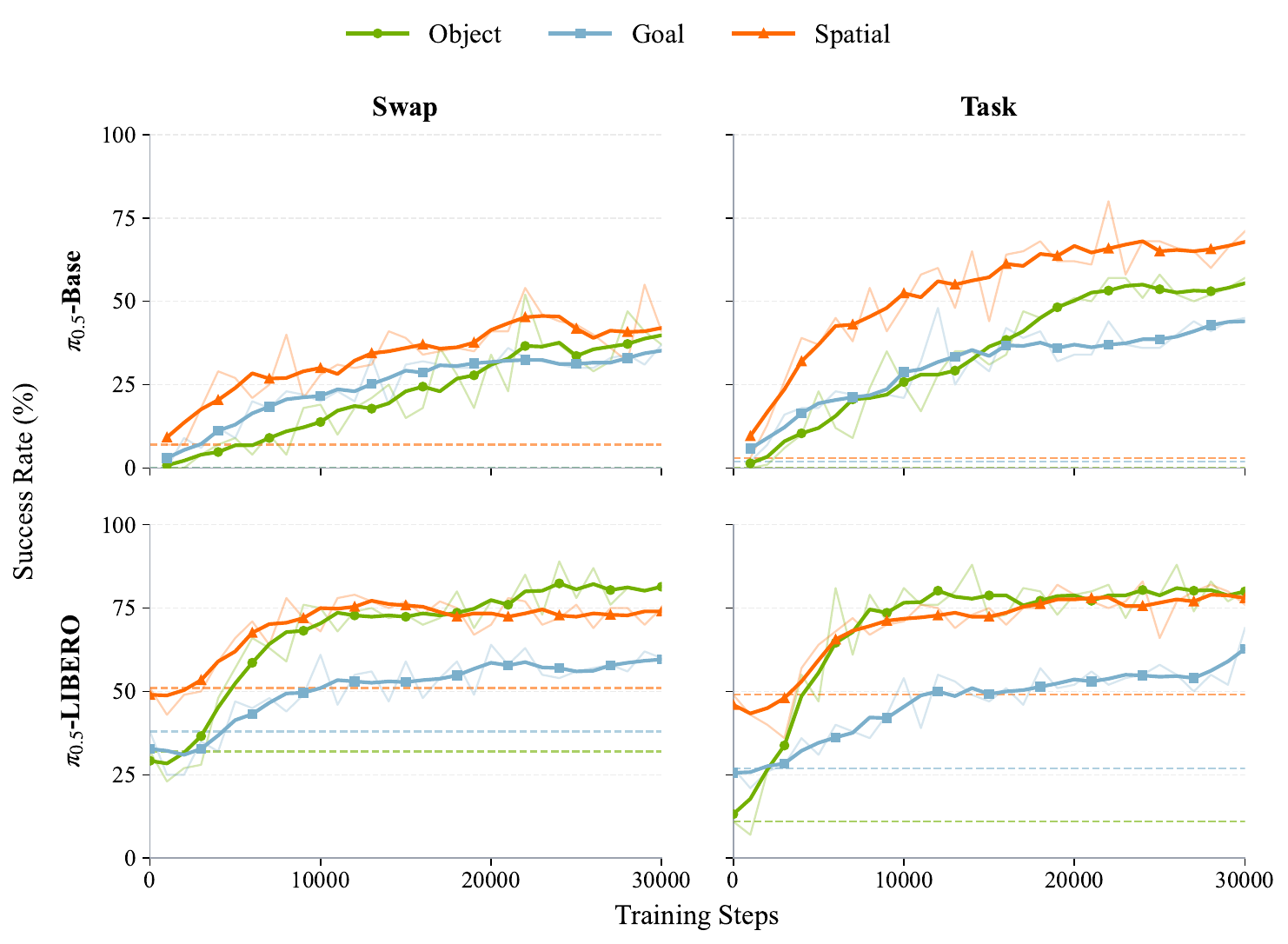}
\caption{VLA fine-tuning on the full harness-trajectory mixture. Both $\pi_{0.5}$-Base and $\pi_{0.5}$-LIBERO use trajectories from all three suites under the Base, Swap, and Task perturbations, sampling the nine categories equally. Curves show test success rate (\%) against fine-tuning steps under Swap and Task perturbations. Faint lines show raw checkpoint results, and solid curves show five-checkpoint centered moving averages, with markers every third checkpoint. Dashed horizontal lines mark the original policy's pre-fine-tuning success rate for each suite and perturbation. For each task, seeds 0-39 form the training set and seeds 40-49 the test set.}
\label{fig:harness-post-training}
\end{figure}

\subsubsection{VLA Fine-Tuning}
\label{sec:harness-post-training-vla}

We first fine-tune $\pi_{0.5}$-LIBERO separately on each suite (Object, Goal, and Spatial), using harness trajectories collected from that suite alone.
Within each suite, we re-run the same converged program under the Base, Swap, and Task conditions and mix the collected trajectories equally (1:1:1).
We evaluate each fine-tuned model on its corresponding suite under the Swap and Task perturbations.
On each task, seeds 0-39 form the training set and seeds 40-49 the test set.

Fig.~\ref{fig:harness-post-training-suites} reports the per-suite result.
We report each suite at its best checkpoint within the 15{,}000-step training budget.
Success increases from 21.5\% to 88.0\% on Object, from 50.0\% to 82.5\% on Spatial, and from 32.5\% to 54.5\% on Goal.
Harness trajectories therefore improve the VLA on all three evaluated suites.

We next fine-tune two starting checkpoints: $\pi_{0.5}$-Base, before any LIBERO fine-tuning, and $\pi_{0.5}$-LIBERO, the official LIBERO-fine-tuned model~\citep{pi05}.
Both use the full mixture of harness trajectories from all three suites under the Base, Swap, and Task conditions.
The resulting nine categories are sampled in equal proportion during fine-tuning.
We evaluate uniformly across the three suites and the two LIBERO-PRO perturbations, covering 60 tasks in total~\citep{liberopro}.
For each task, seeds 0-39 supply the additional fine-tuning data, while seeds 40-49 are held out for evaluation.

Fig.~\ref{fig:harness-post-training} reports success rate against training steps.
We report each starting checkpoint at its best fine-tuning step within the 30{,}000-step training budget.
Fine-tuning increases LIBERO-PRO success from 34.9\% to 73.7\% for $\pi_{0.5}$-LIBERO, a gain of 38.8 percentage points.
For $\pi_{0.5}$-Base, success increases from near zero to 53.3\% using harness trajectories as the additional fine-tuning data.
Both starting checkpoints improve across all three suites and both perturbations.
These results show that the collected demonstrations improve robustness beyond the original checkpoints, including a model without prior LIBERO fine-tuning.

\subsubsection{WAM Fine-Tuning}
\label{sec:harness-post-training-wam}

\begin{wrapfigure}{r}{0.5\textwidth}
\vspace{-1em}
\centering
\includegraphics[width=\linewidth]{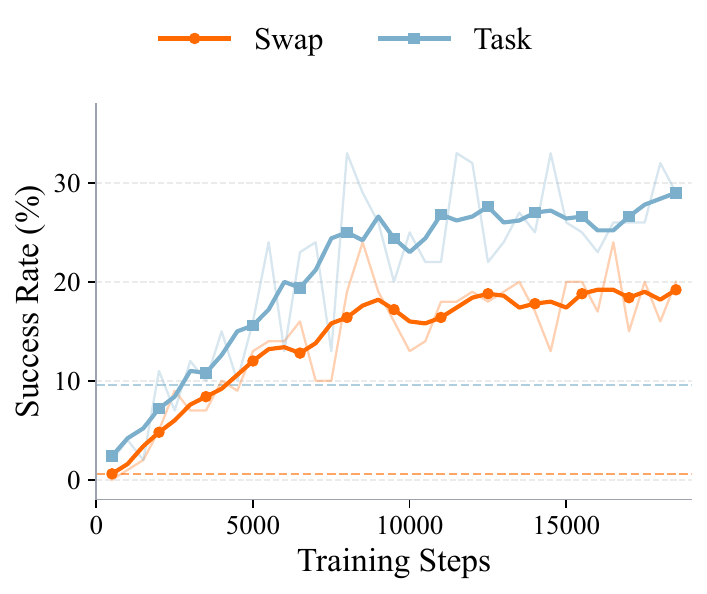}
\caption{WAM fine-tuning on harness trajectories. Curves show DreamZero success rate (\%) on the LIBERO-Object Swap and Task perturbations against fine-tuning steps. Faint lines show raw checkpoint results, and solid curves show five-checkpoint centered moving averages, with markers every third checkpoint. Dashed lines mark pre-fine-tuning success: 0.6\% on Swap and 9.6\% on Task. Checkpoint selection and reported scores use the raw results.}
\label{fig:dreamzero-post-training}
\vspace{-1em}
\end{wrapfigure}

We next fine-tune DreamZero on harness-generated trajectories and evaluate the LIBERO-Object Swap and Task perturbations.
We use the same LIBERO-Object collection protocol as the per-suite $\pi_{0.5}$ experiment, mixing Base, Swap, and Task trajectories equally (1:1:1).
For each task, seeds 0-39 supply fine-tuning trajectories and seeds 40-49 are held out for evaluation.
Before fine-tuning, DreamZero achieves 0.6\% success on Swap and 9.6\% on Task, giving an average success rate of 5.1\%.
We select a single checkpoint by the highest mean of its raw Swap and Task success rates among all observed checkpoints.
Fig.~\ref{fig:dreamzero-post-training} shows that harness-based fine-tuning improves DreamZero on both perturbations.
The best shared checkpoint occurs at 8{,}500 steps, achieving 24.0\% on Swap and 29.0\% on Task, with a mean of 26.5\%.
The gains over the original model are 23.4 points on Swap and 19.4 points on Task, or 21.4 points on average.
Success remains below that of the harness wrapped around DreamZero, but the collected trajectories also improve the standalone WAM through supervised fine-tuning.
Across the two studies, the converged harness serves both as an execution framework and as a source of demonstrations for improving VLA and WAM parameters.
\FloatBarrier

%% file: chapters/6_discussion.tex
\section{Discussion}
\label{sec:discussion}

Three questions frame what \CodePhysical{} does and does not do: why a hybrid of code and a frozen action model outperforms either of its ingredients alone, under what conditions that advantage holds, and where the demonstrations that seed the loop can come from.
We take each in turn.

\subsection{Why the Framework Works}
\label{sec:disc-why}

\CodePhysical{}'s advantage is best read against the two extremes it refuses to occupy.
A pure VLA is built around a single component, the action model, and the recipe that trains it quietly erodes the other two.
The vision-language backbone that supplies semantic perception is overwritten by an action-prediction objective, so the resulting policy keeps its physical competence (the \emph{action}) while its grounding (the \emph{vision}) and its instruction understanding (the \emph{language}) degrade; and its generative action head is reliable at coarse, low-frequency motion but unstable at the fine displacements that delicate manipulation demands.
Section~\ref{sec:why-harness} isolated each failure: $\pi_{0.5}$ grasps the memorized bowl despite a negated instruction, reaches for the old plate position after two objects are swapped, and a WAM's sampled press misses a small button.
A pure code-as-policy system occupies the opposite extreme and fails for the mirror-image reason.
Code supplies grounding, geometry, and verification, but it cannot manufacture contact-rich dexterity: the fine, high-frequency motions a learned action head produces cheaply are precisely the ones hardest to write as explicit statements, which is why a code-only agent collapses to an 18.3\% success rate on LIBERO (Table~\ref{tab:libero}).

\CodePhysical{} sits at the intersection.
It asks code to do what code does well, perceive \emph{what is where}, re-ground a referring expression as an explicit predicate, reason geometrically, and verify each step before committing it, and it asks the frozen action model to do what only it does well: the atomic physical acts, delegated only at the phases the model is competent at.
Neither ingredient is asked to substitute for the other.
Code restores the vision and language the VLA lost and checks the action it retained, without asking code to reproduce that action from scratch; the model supplies the physical competence that no program can express, without being trusted to re-derive the scene or the instruction.
The gain over each extreme is therefore the same gap, viewed from both sides: the framework replaces the degraded V and L of the VLA with code, and replaces the missing A of code with the VLA.

\subsection{When the Framework Works}
\label{sec:disc-when}

The framework is not a universal repair.
Its value is conditional on a property of the underlying model that the previous subsection took for granted: a VLA is not purely end-to-end, and retains \emph{atomic} competence, the ability to grasp a localized object or execute a short reach, even when it can no longer chain those primitives into a long-horizon plan or re-ground them under a perturbation.
This is the property \CodePhysical{} exploits.
The capability attribution of Section~\ref{sec:evolution} labels each phase \emph{good}, \emph{needs code}, or \emph{mixed}, and writes code only where the model is weak, delegating the competent atomic phases straight back to the VLA.
The harness therefore organizes, grounds, and verifies, but it does not manufacture physical competence that is absent.
When a model's primitive abilities themselves fail, when it cannot grasp at all, rather than merely grasping the wrong object, no amount of code repair helps, because there is no competent action for the code to organize.

This boundary explains the one place the framework's gains are comparatively small.
On RoboCasa's composite tasks (Table~\ref{tab:robocasa}) the underlying WorldDreamer executes the middle of a long-horizon trajectory opaquely, and its intermediate atomic steps are both harder for code to inspect and weaker to begin with; \CodePhysical{} corrects the frame of the task but cannot steer the parts it cannot see.
The precondition is the same in both directions: the framework improves a model in proportion to the atomic competence that model already has, and leaves untouched whatever the model fundamentally cannot do.

\subsection{Where the Expert Data Comes From}
\label{sec:disc-data}

The task graph of Section~\ref{sec:evolution} is distilled from an expert demonstration, which might seem to chain the framework to human teleoperation.
It does not.
What the graph actually requires is a \emph{skeleton}, which phases exist, and in what order, not a polished trajectory, and a skeleton can be read off far weaker data.
A single successful \emph{video}, one run that happened to reach the goal, even by chance, is enough to recover that skeleton, because the graph keeps only the phase structure and discards the imperfect motion that produced it.
Data from a \emph{similar} task serves the same role: the graph and the skill library are organized by node, so a grasp or transport phase learned on one task transfers to a novel task that shares the node (Section~\ref{sec:skills}), and a related task's demonstration can therefore seed a new one it was never written for.
Neither source demands a human expert in the loop.

Once the harness has converged, this reliance shrinks further.
As Section~\ref{sec:model-feedback} describes, a converged code policy is itself a trajectory generator, and its successful rollouts, recorded in exactly the scenes where human demonstrations are absent, become the training data for the action model.
The framework's demand for expert data is thus weakest precisely where the supply is rarest: it needs a human label only to sketch a graph, and can bootstrap everything after that from occasional successes, related tasks, and its own converged rollouts.
Section~\ref{sec:harness-post-training} demonstrates this benefit for both the $\pi_{0.5}$ VLA and the DreamZero WAM.

%% file: chapters/7_conclusion.tex
\section{Conclusion}
\label{sec:conclusion}

This paper argued that Physical AI has been organized around a single component, the action model, and that this action-model-centric bias lets the agent's perception, grounding, and planning atrophy inside an end-to-end policy.
We proposed \CodePhysical{}, a code framework that treats code not as a replacement for a VLA or WAM but as the executable harness that organizes perception, geometric reasoning, and action-model invocation.
The framework separates two timescales: within a rollout it executes \emph{open-loop}, grounding each decision and verifying each step before the world can be harmed; across rollouts it evolves \emph{closed-loop}, feeding execution feedback back to revise the program and distilling each failure into a node-keyed skill that transfers to later tasks.
Once the harness converges, its own successful trajectories are fed back to train the action model, closing the gap on tasks for which no expert demonstration exists.

We evaluated \CodePhysical{} across robot arms, a humanoid, and a mobile robot, driven by three generations of action interface, GraspNet, $\pi_{0.5}$, and WorldDreamer, together with a pure API containing no learned model.
On LIBERO it reaches 98.1\% overall, a new state of the art, and on LIBERO-PRO it attains 96.5\% by recovering from the perturbations that collapse fixed policies.
On RoboCasa it lifts WorldDreamer from 65.0\% to 92.2\% on atomic tasks.
Because open-loop execution puts no model call on the critical path, these gains come nearly free at rollout time, orders of magnitude faster and without per-run API cost.

Taken together, the results indicate that the frontier of Physical AI lies not only in the action model but in the executable system that organizes, checks, and evolves it.
Code, used as an interface rather than a substitute for learned control, is a practical path to more robust and more inspectable physical agents.

\section{Future Work}
\label{sec:future-work}

\CodePhysical{} inherits much of its competence from the action model it wraps, so a natural direction for future work is to make that dependency as thin and uniform as an API. Whether the underlying model is a PPO-trained policy, a VLA, or a WAM of any paradigm, the ideal interface is a single contract: given an instruction, the model reliably completes one atomic action. Standardizing this contract would let the harness compose any action model without reasoning about its internals, and would let \CodePhysical{} benefit directly from progress across all model families.

A second direction is to reduce our reliance on online perception. The code depends heavily on perception to ground each decision, which is both a source of fragility and a recurring cost during evolution. Building a map or a scene model ahead of time, or closing the sim-to-real gap through direct real-to-sim transfer, could offload much of this burden, making perception both cheaper and more stable.

Finally, our method is not the opposite of directly invoking a coding agent; it is complementary to one. A coding agent can supply the open-ended reasoning that a fixed program cannot, while \CodePhysical{} supplies validated, reusable code packages. Combined, an agent completing a long-horizon task need not deliberate or invoke a model at every step; it can instead reuse or lightly adapt a code package of similar functionality, reserving its reasoning for genuinely novel situations.

%% file: chapters/acknowledgments.tex
\phantomsection
\section*{Contributions and Acknowledgments}
\label{sec:contributions-acknowledgments}
\addcontentsline{toc}{section}{Contributions and Acknowledgments}

\noindent
\begin{minipage}[t]{0.47\textwidth}
\paragraph{Core Contributors.}
\begin{itemize}
    \item Xin Wang$^{1,2,*}$
    \item Wenhao Wu$^{1,3,*}$
    \item Menghao Zhang$^{1,*}$
    \item Zhi Wang$^{3,\dagger}$
    \item Kun Shao$^{1,\dagger, \ddagger}$
    \item Jian Luan$^{1,\dagger}$
\end{itemize}
\end{minipage}\hfill
\begin{minipage}[t]{0.47\textwidth}
\paragraph{Contributors.}
\begin{itemize}
    \item Yang Li$^{4}$
    \item Qing Li$^{1}$
    \item Shangding Gu$^{4}$
    \item Huichi Zhou$^{6}$
    \item Shuqing Shi$^{7}$
    \item Fei Ni$^{8}$
    \item Shuo Lu$^{9}$
    \item Weicheng Meng$^{10}$
    \item Kang Li$^{11}$
    \item Jin Wu$^{5}$
    \item Kang Zhao$^{1}$
    \item Shangmin Guo$^{12}$
    \item Gen Li$^{13}$
    \item Yongqiang Tang$^{9}$
    \item Zhizhong Zhang$^{5}$
    \item Yuan Xie$^{5}$
    \item Heng Qu$^{1}$
\end{itemize}
\end{minipage}

\par\smallskip
{\footnotesize $^{*}$ Equal contribution.
\textsuperscript{\textdagger} Corresponding author.
\textsuperscript{$\ddagger$} Project lead.\par
\smallskip
$^{1}$Xiaomi Inc., $^{2}$Tsinghua University, $^{3}$Nanjing University, $^{4}$Shanghai Jiao Tong University, $^{5}$East China Normal University, $^{6}$University College London, $^{7}$King's College London, $^{8}$Imperial College London, $^{9}$Institute of Automation, Chinese Academy of Sciences, $^{10}$Shanghai Innovation Institute, $^{11}$University of Oxford, $^{12}$University of Edinburgh, $^{13}$Nanyang Technological University.}

%% file: chapters/appendix.tex
\clearpage
\beginappendix \appendix
\section{Main Benchmark Results Table}
\label{app:main-results}

Table~\ref{tab:robosuite-per-task} reports the per-task success rate on robosuite underlying Fig.~\ref{fig:robosuite-per-task}.
The final row gives the improvement of \CodePhysical{} over ASPIRE.

\begin{table}[htbp]
\centering
\caption{Per-task success rate on robosuite (\%).}
\label{tab:robosuite-per-task}
\begin{tabular}{lcccccccc}
\toprule
Method & 2A-Hand & 2A-Lift & Insert & Lift & Stack & Restack & Wipe & Avg \\
\midrule
CaP-Agent & 20 & 74 & 0 & \cellcolor{miSecond}97 & 98 & \cellcolor{miSecond}89 & \cellcolor{miBest}100 & 68.3 \\
Human & 91 & \cellcolor{miBest}94 & \cellcolor{miSecond}80 & 93 & 73 & 47 & \cellcolor{miBest}100 & \cellcolor{miSecond}82.6 \\
ASPIRE & \cellcolor{miSecond}92 & 71 & 9 & \cellcolor{miSecond}97 & \cellcolor{miSecond}99 & \cellcolor{miBest}100 & \cellcolor{miSecond}99 & 81.0 \\
\CodePhysical{} & \cellcolor{miBest}100 & \cellcolor{miSecond}91 & \cellcolor{miBest}83 & \cellcolor{miBest}100 & \cellcolor{miBest}100 & \cellcolor{miBest}100 & \cellcolor{miBest}100 & \cellcolor{miBest}96.3 \\
\midrule
$\Delta$ vs.\ ASPIRE & \textcolor{miBest}{+8.0} & \textcolor{miBest}{+20.0} & \textcolor{miBest}{+74.0} & \textcolor{miBest}{+3.0} & \textcolor{miBest}{+1.0} & \textcolor{miBest}{0.0} & \textcolor{miBest}{+1.0} & \textcolor{miBest}{+15.3} \\
\bottomrule
\end{tabular}
\end{table}

Table~\ref{tab:libero-pro-reproduction} reports the full LIBERO-PRO comparison underlying Fig.~\ref{fig:libero-reproduction}, aggregated by suite and perturbation across all eight baselines.
The final row gives the improvement of \CodePhysical{} over the strongest pure VLA, $\pi_{0.5}$.

\begin{table}[htbp]
\centering
\caption{LIBERO-PRO success rate by suite and perturbation (\%).}
\label{tab:libero-pro-reproduction}
\begin{tabular}{lccccccc}
\toprule
& \multicolumn{2}{c}{Object} & \multicolumn{2}{c}{Goal} & \multicolumn{2}{c}{Spatial} & \\
\cmidrule(lr){2-3} \cmidrule(lr){4-5} \cmidrule(lr){6-7}
Method & Swap & Task & Swap & Task & Swap & Task & Avg \\
\midrule
OpenVLA & 0 & 0 & 0 & 0 & 0 & 0 & 0.0 \\
$\pi_0$ & 0 & 0 & 0 & 0 & 0 & 0 & 0.0 \\
$\pi_{0.5}$ & 29.8 & 10.8 & 37.6 & 29.0 & 51.8 & 50.6 & 34.9 \\
CaP-Agent & 22 & 18 & 26 & 17 & 12 & 14 & 18.2 \\
ASPIRE & \cellcolor{miSecond}98 & \cellcolor{miSecond}95 & 81 & 45 & 51 & 60 & 71.7 \\
Harness VLA (Codex) & 91 & 94 & 66 & 75 & 69 & 81 & 79.3 \\
Harness VLA (CC) & 90 & 88 & \cellcolor{miSecond}87 & \cellcolor{miSecond}87 & \cellcolor{miSecond}80 & \cellcolor{miSecond}94 & \cellcolor{miSecond}87.7 \\
\CodePhysical{} & \cellcolor{miBest}98.6 & \cellcolor{miBest}99.2 & \cellcolor{miBest}94.6 & \cellcolor{miBest}97.8 & \cellcolor{miBest}93.8 & \cellcolor{miBest}95.2 & \cellcolor{miBest}96.5 \\
\midrule
$\Delta$ vs.\ $\pi_{0.5}$ & \textcolor{miBest}{+68.8} & \textcolor{miBest}{+88.4} & \textcolor{miBest}{+57.0} & \textcolor{miBest}{+68.8} & \textcolor{miBest}{+42.0} & \textcolor{miBest}{+44.6} & \textcolor{miBest}{+61.6} \\
\bottomrule
\end{tabular}
\end{table}

\section{Detailed Results}
\label{app:details}

\subsection{LIBERO}
\label{app:libero-details}

Table~\ref{tab:libero-per-task} reports the per-task results of our method on the three LIBERO suites, which underlie the Spatial, Object, and Goal columns in Table~\ref{tab:libero}.

Table~\ref{tab:pi05-libero-per-task} reports the per-task results of the underlying $\pi_{0.5}$-LIBERO model on the three LIBERO suites, which serve as the baseline against which the \CodePhysical{} results in Table~\ref{tab:libero-per-task} are measured.

Table~\ref{tab:dreamzero-libero-per-task} reports the per-task results of DreamZero-LIBERO, our own evaluation of the World Action Model DreamZero on standard LIBERO, which underlie the DreamZero-LIBERO row in Table~\ref{tab:libero}.
\begin{table}[htbp]
\centering
\caption{Per-task results of \CodePhysical  on the three LIBERO suites (success rate, \%). All results are our own evaluation over 50 seeds per task.}
\label{tab:libero-per-task}
{\small
\begin{tabular}{lp{8.5cm}cc}
\toprule
Task & Description & Success / Total & Success Rate \\
\midrule
\multicolumn{4}{l}{\textbf{LIBERO-Object}} \\
\midrule
Task 0 & Pick up the alphabet soup and place it in the basket & 50/50 & 100.0 \\
Task 1 & Pick up the cream cheese and place it in the basket & 50/50 & 100.0 \\
Task 2 & Pick up the salad dressing and place it in the basket & 50/50 & 100.0 \\
Task 3 & Pick up the BBQ sauce and place it in the basket & 50/50 & 100.0 \\
Task 4 & Pick up the ketchup and place it in the basket & 50/50 & 100.0 \\
Task 5 & Pick up the tomato sauce and place it in the basket & 50/50 & 100.0 \\
Task 6 & Pick up the butter and place it in the basket & 50/50 & 100.0 \\
Task 7 & Pick up the milk and place it in the basket & 50/50 & 100.0 \\
Task 8 & Pick up the chocolate pudding and place it in the basket & 50/50 & 100.0 \\
Task 9 & Pick up the orange juice and place it in the basket & 50/50 & 100.0 \\
\midrule
\multicolumn{2}{l}{\textbf{Object subtotal}} & 500/500 & \textbf{100.0} \\
\midrule
\multicolumn{4}{l}{\textbf{LIBERO-Spatial}} \\
\midrule
Task 0 & Pick up the black bowl between the plate and the ramekin and place it on the plate & 50/50 & 100.0 \\
Task 1 & Pick up the black bowl next to the ramekin and place it on the plate & 50/50 & 100.0 \\
Task 2 & Pick up the black bowl from table center and place it on the plate & 50/50 & 100.0 \\
Task 3 & Pick up the black bowl on the cookie box and place it on the plate & 48/50 & 96.0 \\
Task 4 & Pick up the black bowl in the top drawer of the wooden cabinet and place it on the plate & 44/50 & 88.0 \\
Task 5 & Pick up the black bowl on the ramekin and place it on the plate & 48/50 & 96.0 \\
Task 6 & Pick up the black bowl next to the cookie box and place it on the plate & 50/50 & 100.0 \\
Task 7 & Pick up the black bowl on the stove and place it on the plate & 47/50 & 94.0 \\
Task 8 & Pick up the black bowl next to the plate and place it on the plate & 50/50 & 100.0 \\
Task 9 & Pick up the black bowl on the wooden cabinet and place it on the plate & 49/50 & 98.0 \\
\midrule
\multicolumn{2}{l}{\textbf{Spatial subtotal}} & 486/500 & \textbf{97.2} \\
\midrule
\multicolumn{4}{l}{\textbf{LIBERO-Goal}} \\
\midrule
Task 0 & Open the middle drawer of the cabinet & 46/50 & 92.0 \\
Task 1 & Put the bowl on the stove & 50/50 & 100.0 \\
Task 2 & Put the wine bottle on top of the cabinet & 49/50 & 98.0 \\
Task 3 & Open the top drawer and put the bowl inside & 44/50 & 88.0 \\
Task 4 & Put the bowl on top of the cabinet & 48/50 & 96.0 \\
Task 5 & Push the plate to the front of the stove & 50/50 & 100.0 \\
Task 6 & Put the cream cheese in the bowl & 50/50 & 100.0 \\
Task 7 & Turn on the stove & 50/50 & 100.0 \\
Task 8 & Put the bowl on the plate & 49/50 & 98.0 \\
Task 9 & Put the wine bottle on the rack & 50/50 & 100.0 \\
\midrule
\multicolumn{2}{l}{\textbf{Goal subtotal}} & 486/500 & \textbf{97.2} \\
\midrule
\multicolumn{2}{l}{\textbf{Overall}} & 1472/1500 & \textbf{98.1} \\
\bottomrule
\end{tabular}
}
\end{table}

\begin{table}[htbp]
\centering
\caption{Per-task results of $\pi_{0.5}$ on the three LIBERO suites (success rate, \%).  All results are our own evaluation over 50 seeds per task.}
\label{tab:pi05-libero-per-task}
{\small
\begin{tabular}{lp{8.5cm}cc}
\toprule
Task & Description & Success / Total & Success Rate \\
\midrule
\multicolumn{4}{l}{\textbf{LIBERO-Object}} \\
\midrule
Task 0 & Pick up the alphabet soup and place it in the basket & 50/50 & 100.0 \\
Task 1 & Pick up the cream cheese and place it in the basket & 50/50 & 100.0 \\
Task 2 & Pick up the salad dressing and place it in the basket & 50/50 & 100.0 \\
Task 3 & Pick up the BBQ sauce and place it in the basket & 50/50 & 100.0 \\
Task 4 & Pick up the ketchup and place it in the basket & 50/50 & 100.0 \\
Task 5 & Pick up the tomato sauce and place it in the basket & 50/50 & 100.0 \\
Task 6 & Pick up the butter and place it in the basket & 50/50 & 100.0 \\
Task 7 & Pick up the milk and place it in the basket & 50/50 & 100.0 \\
Task 8 & Pick up the chocolate pudding and place it in the basket & 50/50 & 100.0 \\
Task 9 & Pick up the orange juice and place it in the basket & 50/50 & 100.0 \\
\midrule
\multicolumn{2}{l}{\textbf{Object subtotal}} & 500/500 & \textbf{100.0} \\
\midrule
\multicolumn{4}{l}{\textbf{LIBERO-Spatial}} \\
\midrule
Task 0 & Pick up the black bowl between the plate and the ramekin and place it on the plate & 50/50 & 100.0 \\
Task 1 & Pick up the black bowl next to the ramekin and place it on the plate & 50/50 & 100.0 \\
Task 2 & Pick up the black bowl from table center and place it on the plate & 49/50 & 98.0 \\
Task 3 & Pick up the black bowl on the cookie box and place it on the plate & 48/50 & 96.0 \\
Task 4 & Pick up the black bowl in the top drawer of the wooden cabinet and place it on the plate & 34/50 & 68.0 \\
Task 5 & Pick up the black bowl on the ramekin and place it on the plate & 50/50 & 100.0 \\
Task 6 & Pick up the black bowl next to the cookie box and place it on the plate & 50/50 & 100.0 \\
Task 7 & Pick up the black bowl on the stove and place it on the plate & 48/50 & 96.0 \\
Task 8 & Pick up the black bowl next to the plate and place it on the plate & 50/50 & 100.0 \\
Task 9 & Pick up the black bowl on the wooden cabinet and place it on the plate & 49/50 & 98.0 \\
\midrule
\multicolumn{2}{l}{\textbf{Spatial subtotal}} & 478/500 & \textbf{95.6} \\
\midrule
\multicolumn{4}{l}{\textbf{LIBERO-Goal}} \\
\midrule
Task 0 & Open the middle drawer of the cabinet & 36/50 & 72.0 \\
Task 1 & Put the bowl on the stove & 50/50 & 100.0 \\
Task 2 & Put the wine bottle on top of the cabinet & 50/50 & 100.0 \\
Task 3 & Open the top drawer and put the bowl inside & 46/50 & 92.0 \\
Task 4 & Put the bowl on top of the cabinet & 47/50 & 94.0 \\
Task 5 & Push the plate to the front of the stove & 50/50 & 100.0 \\
Task 6 & Put the cream cheese in the bowl & 50/50 & 100.0 \\
Task 7 & Turn on the stove & 50/50 & 100.0 \\
Task 8 & Put the bowl on the plate & 48/50 & 96.0 \\
Task 9 & Put the wine bottle on the rack & 49/50 & 98.0 \\
\midrule
\multicolumn{2}{l}{\textbf{Goal subtotal}} & 476/500 & \textbf{95.2} \\
\midrule
\multicolumn{2}{l}{\textbf{Overall}} & 1454/1500 & \textbf{96.9} \\
\bottomrule
\end{tabular}
}
\end{table}

\begin{table}[htbp]
\centering
\caption{Per-task results of DreamZero-LIBERO on the three LIBERO suites (success rate, \%). All results are our own evaluation over 50 seeds per task.}
\label{tab:dreamzero-libero-per-task}
{\small
\begin{tabular}{lp{8.5cm}cc}
\toprule
Task & Description & Success / Total & Success Rate \\
\midrule
\multicolumn{4}{l}{\textbf{LIBERO-Object}} \\
\midrule
Task 0 & Pick up the alphabet soup and place it in the basket & 45/50 & 90.0 \\
Task 1 & Pick up the cream cheese and place it in the basket & 50/50 & 100.0 \\
Task 2 & Pick up the salad dressing and place it in the basket & 50/50 & 100.0 \\
Task 3 & Pick up the BBQ sauce and place it in the basket & 46/50 & 92.0 \\
Task 4 & Pick up the ketchup and place it in the basket & 41/50 & 82.0 \\
Task 5 & Pick up the tomato sauce and place it in the basket & 50/50 & 100.0 \\
Task 6 & Pick up the butter and place it in the basket & 46/50 & 92.0 \\
Task 7 & Pick up the milk and place it in the basket & 28/50 & 56.0 \\
Task 8 & Pick up the chocolate pudding and place it in the basket & 48/50 & 96.0 \\
Task 9 & Pick up the orange juice and place it in the basket & 47/50 & 94.0 \\
\midrule
\multicolumn{2}{l}{\textbf{Object subtotal}} & 451/500 & \textbf{90.2} \\
\midrule
\multicolumn{4}{l}{\textbf{LIBERO-Spatial}} \\
\midrule
Task 0 & Pick up the black bowl between the plate and the ramekin and place it on the plate & 29/50 & 58.0 \\
Task 1 & Pick up the black bowl next to the ramekin and place it on the plate & 43/50 & 86.0 \\
Task 2 & Pick up the black bowl from table center and place it on the plate & 47/50 & 94.0 \\
Task 3 & Pick up the black bowl on the cookie box and place it on the plate & 48/50 & 96.0 \\
Task 4 & Pick up the black bowl in the top drawer of the wooden cabinet and place it on the plate & 35/50 & 70.0 \\
Task 5 & Pick up the black bowl on the ramekin and place it on the plate & 34/50 & 68.0 \\
Task 6 & Pick up the black bowl next to the cookie box and place it on the plate & 48/50 & 96.0 \\
Task 7 & Pick up the black bowl on the stove and place it on the plate & 33/50 & 66.0 \\
Task 8 & Pick up the black bowl next to the plate and place it on the plate & 42/50 & 84.0 \\
Task 9 & Pick up the black bowl on the wooden cabinet and place it on the plate & 49/50 & 98.0 \\
\midrule
\multicolumn{2}{l}{\textbf{Spatial subtotal}} & 408/500 & \textbf{81.6} \\
\midrule
\multicolumn{4}{l}{\textbf{LIBERO-Goal}} \\
\midrule
Task 0 & Open the middle drawer of the cabinet & 24/50 & 48.0 \\
Task 1 & Put the bowl on the stove & 44/50 & 88.0 \\
Task 2 & Put the wine bottle on top of the cabinet & 16/50 & 32.0 \\
Task 3 & Open the top drawer and put the bowl inside & 7/50 & 14.0 \\
Task 4 & Put the bowl on top of the cabinet & 0/50 & 0.0 \\
Task 5 & Push the plate to the front of the stove & 13/50 & 26.0 \\
Task 6 & Put the cream cheese in the bowl & 12/50 & 24.0 \\
Task 7 & Turn on the stove & 45/50 & 90.0 \\
Task 8 & Put the bowl on the plate & 43/50 & 86.0 \\
Task 9 & Put the wine bottle on the rack & 2/50 & 4.0 \\
\midrule
\multicolumn{2}{l}{\textbf{Goal subtotal}} & 206/500 & \textbf{41.2} \\
\midrule
\multicolumn{2}{l}{\textbf{Overall}} & 1065/1500 & \textbf{71.0} \\
\bottomrule
\end{tabular}
}
\end{table}

\subsection{LIBERO-PRO Perturbations}
\label{app:libero-pro-perturb}

Table~\ref{tab:libero-perturb} reports the per-task breakdown of \CodePhysical{} under the two LIBERO-PRO perturbation axes we evaluate.
Table~\ref{tab:pi05-libero-perturb} reports the per-task results of the underlying $\pi_{0.5}$ model on LIBERO-PRO under the Swap and Task perturbations, the baseline against which the \CodePhysical{} results in Table~\ref{tab:libero-perturb} are measured.

\begin{table}[htbp]
\centering
\caption{Per-task results of \CodePhysical on LIBERO-PRO under Swap and Task perturbations (success rate, \%).}
\label{tab:libero-perturb}
\begin{tabular}{lcccccc}
\toprule
& \multicolumn{3}{c}{Swap} & \multicolumn{3}{c}{Task} \\
\cmidrule(lr){2-4} \cmidrule(lr){5-7}
Task & Succ./Total & Fail & Rate & Succ./Total & Fail & Rate \\
\midrule
\multicolumn{7}{l}{\textbf{LIBERO-Object}} \\
\midrule
Task 0 & 50/50 & 0 & 100.0 & 50/50 & 0 & 100.0 \\
Task 1 & 50/50 & 0 & 100.0 & 50/50 & 0 & 100.0 \\
Task 2 & 50/50 & 0 & 100.0 & 47/50 & 3 & 94.0 \\
Task 3 & 50/50 & 0 & 100.0 & 50/50 & 0 & 100.0 \\
Task 4 & 50/50 & 0 & 100.0 & 49/50 & 1 & 98.0 \\
Task 5 & 45/50 & 5 & 90.0 & 50/50 & 0 & 100.0 \\
Task 6 & 50/50 & 0 & 100.0 & 50/50 & 0 & 100.0 \\
Task 7 & 50/50 & 0 & 100.0 & 50/50 & 0 & 100.0 \\
Task 8 & 50/50 & 0 & 100.0 & 50/50 & 0 & 100.0 \\
Task 9 & 48/50 & 2 & 96.0 & 50/50 & 0 & 100.0 \\
\midrule
\multicolumn{1}{l}{\textbf{Object subtotal}} & 493/500 & 7 & \textbf{98.6} & 496/500 & 4 & \textbf{99.2} \\
\midrule
\multicolumn{7}{l}{\textbf{LIBERO-Goal}} \\
\midrule
Task 0 & 44/50 & 6 & 88.0 & 49/50 & 1 & 98.0 \\
Task 1 & 49/50 & 1 & 98.0 & 45/50 & 5 & 90.0 \\
Task 2 & 43/50 & 7 & 86.0 & 47/50 & 3 & 94.0 \\
Task 3 & 48/50 & 2 & 96.0 & 49/50 & 1 & 98.0 \\
Task 4 & 49/50 & 1 & 98.0 & 49/50 & 1 & 98.0 \\
Task 5 & 48/50 & 2 & 96.0 & 50/50 & 0 & 100.0 \\
Task 6 & 47/50 & 3 & 94.0 & 50/50 & 0 & 100.0 \\
Task 7 & 50/50 & 0 & 100.0 & 50/50 & 0 & 100.0 \\
Task 8 & 50/50 & 0 & 100.0 & 50/50 & 0 & 100.0 \\
Task 9 & 45/50 & 5 & 90.0 & 50/50 & 0 & 100.0 \\
\midrule
\multicolumn{1}{l}{\textbf{Goal subtotal}} & 473/500 & 27 & \textbf{94.6} & 489/500 & 11 & \textbf{97.8} \\
\midrule
\multicolumn{7}{l}{\textbf{LIBERO-Spatial}} \\
\midrule
Task 0 & 50/50 & 0 & 100.0 & 46/50 & 4 & 92.0 \\
Task 1 & 50/50 & 0 & 100.0 & 50/50 & 0 & 100.0 \\
Task 2 & 49/50 & 1 & 98.0 & 45/50 & 5 & 90.0 \\
Task 3 & 45/50 & 5 & 90.0 & 45/50 & 5 & 90.0 \\
Task 4 & 46/50 & 4 & 92.0 & 44/50 & 6 & 88.0 \\
Task 5 & 50/50 & 0 & 100.0 & 50/50 & 0 & 100.0 \\
Task 6 & 45/50 & 5 & 90.0 & 49/50 & 1 & 98.0 \\
Task 7 & 39/50 & 11 & 78.0 & 50/50 & 0 & 100.0 \\
Task 8 & 50/50 & 0 & 100.0 & 48/50 & 2 & 96.0 \\
Task 9 & 45/50 & 5 & 90.0 & 49/50 & 1 & 98.0 \\
\midrule
\multicolumn{1}{l}{\textbf{Spatial subtotal}} & 469/500 & 31 & \textbf{93.8} & 476/500 & 24 & \textbf{95.2} \\
\midrule
\multicolumn{1}{l}{\textbf{Overall}} & 1435/1500 & 65 & \textbf{95.7} & 1461/1500 & 39 & \textbf{97.4} \\
\bottomrule
\end{tabular}
\end{table}

\begin{table}[htbp]
\centering
\caption{Per-task results of $\pi_{0.5}$ on LIBERO-PRO under Swap and Task perturbations (success rate, \%).}
\label{tab:pi05-libero-perturb}
\begin{tabular}{lcccccc}
\toprule
& \multicolumn{3}{c}{Swap} & \multicolumn{3}{c}{Task} \\
\cmidrule(lr){2-4} \cmidrule(lr){5-7}
Task & Succ./Total & Fail & Rate & Succ./Total & Fail & Rate \\
\midrule
\multicolumn{7}{l}{\textbf{LIBERO-Object}} \\
\midrule
Task 0 & 41/50 & 9 & 82.0 & 0/50 & 50 & 0.0 \\
Task 1 & 50/50 & 0 & 100.0 & 50/50 & 0 & 100.0 \\
Task 2 & 0/50 & 50 & 0.0 & 0/50 & 50 & 0.0 \\
Task 3 & 5/50 & 45 & 10.0 & 0/50 & 50 & 0.0 \\
Task 4 & 0/50 & 50 & 0.0 & 0/50 & 50 & 0.0 \\
Task 5 & 0/50 & 50 & 0.0 & 0/50 & 50 & 0.0 \\
Task 6 & 44/50 & 6 & 88.0 & 0/50 & 50 & 0.0 \\
Task 7 & 0/50 & 50 & 0.0 & 4/50 & 46 & 8.0 \\
Task 8 & 0/50 & 50 & 0.0 & 0/50 & 50 & 0.0 \\
Task 9 & 9/50 & 41 & 18.0 & 0/50 & 50 & 0.0 \\
\midrule
\multicolumn{1}{l}{\textbf{Object subtotal}} & 149/500 & 351 & \textbf{29.8} & 54/500 & 446 & \textbf{10.8} \\
\midrule
\multicolumn{7}{l}{\textbf{LIBERO-Goal}} \\
\midrule
Task 0 & 0/50 & 50 & 0.0 & 0/50 & 50 & 0.0 \\
Task 1 & 50/50 & 0 & 100.0 & 8/50 & 42 & 16.0 \\
Task 2 & 0/50 & 50 & 0.0 & 15/50 & 35 & 30.0 \\
Task 3 & 50/50 & 0 & 100.0 & 17/50 & 33 & 34.0 \\
Task 4 & 0/50 & 50 & 0.0 & 12/50 & 38 & 24.0 \\
Task 5 & 0/50 & 50 & 0.0 & 6/50 & 44 & 12.0 \\
Task 6 & 0/50 & 50 & 0.0 & 9/50 & 41 & 18.0 \\
Task 7 & 49/50 & 1 & 98.0 & 46/50 & 4 & 92.0 \\
Task 8 & 37/50 & 13 & 74.0 & 32/50 & 18 & 64.0 \\
Task 9 & 2/50 & 48 & 4.0 & 0/50 & 50 & 0.0 \\
\midrule
\multicolumn{1}{l}{\textbf{Goal subtotal}} & 188/500 & 312 & \textbf{37.6} & 145/500 & 355 & \textbf{29.0} \\
\midrule
\multicolumn{7}{l}{\textbf{LIBERO-Spatial}} \\
\midrule
Task 0 & 49/50 & 1 & 98.0 & 0/50 & 50 & 0.0 \\
Task 1 & 25/50 & 25 & 50.0 & 44/50 & 6 & 88.0 \\
Task 2 & 50/50 & 0 & 100.0 & 0/50 & 50 & 0.0 \\
Task 3 & 0/50 & 50 & 0.0 & 3/50 & 47 & 6.0 \\
Task 4 & 41/50 & 9 & 82.0 & 0/50 & 50 & 0.0 \\
Task 5 & 44/50 & 6 & 88.0 & 46/50 & 4 & 92.0 \\
Task 6 & 0/50 & 50 & 0.0 & 48/50 & 2 & 96.0 \\
Task 7 & 0/50 & 50 & 0.0 & 49/50 & 1 & 98.0 \\
Task 8 & 50/50 & 0 & 100.0 & 13/50 & 37 & 26.0 \\
Task 9 & 0/50 & 50 & 0.0 & 50/50 & 0 & 100.0 \\
\midrule
\multicolumn{1}{l}{\textbf{Spatial subtotal}} & 259/500 & 241 & \textbf{51.8} & 253/500 & 247 & \textbf{50.6} \\
\midrule
\multicolumn{1}{l}{\textbf{Overall}} & 596/1500 & 904 & \textbf{39.7} & 452/1500 & 1048 & \textbf{30.1} \\
\bottomrule
\end{tabular}
\end{table}

\subsection{RoboCasa}
\label{app:robocasa-details}

Table~\ref{tab:robocasa-per-task} reports the per-task results on the RoboCasa atomic tasks that underlie the aggregate in Table~\ref{tab:robocasa}.
Table~\ref{tab:robocasa-wam-per-task} reports the per-task results of the underlying WorldDreamer WAM on the same RoboCasa atomic tasks, the baseline against which the \CodePhysical{} results in Table~\ref{tab:robocasa-per-task} are measured.

\begin{table}[htbp]
\centering
\caption{Per-task results on RoboCasa atomic tasks (success rate, \%).}
\label{tab:robocasa-per-task}
\begin{tabular}{llccc}
\toprule
Task & Name & Layout / Style & Success / Total & Success Rate \\
\midrule
Task 0 & CloseBlenderLid & 8 / 8 & 19/20 & 95.0 \\
Task 1 & CloseFridge & 2 / 2 & 20/20 & 100.0 \\
Task 2 & CloseToasterOvenDoor & 7 / 7 & 19/20 & 95.0 \\
Task 3 & CoffeeSetupMug & 4 / 4 & 16/20 & 80.0 \\
Task 4 & NavigateKitchen & 5 / 5 & 20/20 & 100.0 \\
Task 5 & OpenCabinet & 7 / 7 & 18/20 & 90.0 \\
Task 6 & OpenDrawer & 7 / 7 & 20/20 & 100.0 \\
Task 7 & OpenStandMixerHead & 2 / 2 & 19/20 & 95.0 \\
Task 8 & PickPlaceCounterToCabinet & 2 / 2 & 19/20 & 95.0 \\
Task 9 & PickPlaceCounterToStove & 6 / 6 & 16/20 & 80.0 \\
Task 10 & PickPlaceDrawerToCounter & 10 / 10 & 15/20 & 75.0 \\
Task 11 & PickPlaceSinkToCounter & 5 / 5 & 17/20 & 85.0 \\
Task 12 & PickPlaceToasterToCounter & 5 / 5 & 20/20 & 100.0 \\
Task 13 & SlideDishwasherRack & 2 / 2 & 19/20 & 95.0 \\
Task 14 & TurnOffStove & 2 / 2 & 18/20 & 90.0 \\
Task 15 & TurnOnElectricKettle & 3 / 3 & 18/20 & 90.0 \\
Task 16 & TurnOnMicrowave & 2 / 2 & 20/20 & 100.0 \\
Task 17 & TurnOnSinkFaucet & 2 / 2 & 19/20 & 95.0 \\
\midrule
\multicolumn{3}{l}{\textbf{Total}} & 332/360 & \textbf{92.2} \\
\bottomrule
\end{tabular}
\end{table}

\begin{table}[htbp]
\centering
\caption{Per-task results of the underlying WorldDreamer WAM on RoboCasa atomic tasks (success rate, \%).}
\label{tab:robocasa-wam-per-task}
\begin{tabular}{llccc}
\toprule
Task & Name & Layout / Style & Success / Total & Success Rate \\
\midrule
Task 0 & CloseBlenderLid & 8 / 8 & 6/20 & 30.0 \\
Task 1 & CloseFridge & 2 / 2 & 20/20 & 100.0 \\
Task 2 & CloseToasterOvenDoor & 7 / 7 & 8/20 & 40.0 \\
Task 3 & CoffeeSetupMug & 4 / 4 & 8/20 & 40.0 \\
Task 4 & NavigateKitchen & 5 / 5 & 8/20 & 40.0 \\
Task 5 & OpenCabinet & 7 / 7 & 18/20 & 90.0 \\
Task 6 & OpenDrawer & 7 / 7 & 14/20 & 70.0 \\
Task 7 & OpenStandMixerHead & 2 / 2 & 20/20 & 100.0 \\
Task 8 & PickPlaceCounterToCabinet & 2 / 2 & 20/20 & 100.0 \\
Task 9 & PickPlaceCounterToStove & 6 / 6 & 16/20 & 80.0 \\
Task 10 & PickPlaceDrawerToCounter & 10 / 10 & 15/20 & 75.0 \\
Task 11 & PickPlaceSinkToCounter & 5 / 5 & 20/20 & 100.0 \\
Task 12 & PickPlaceToasterToCounter & 5 / 5 & 18/20 & 90.0 \\
Task 13 & SlideDishwasherRack & 2 / 2 & 13/20 & 65.0 \\
Task 14 & TurnOffStove & 2 / 2 & 0/20 & 0.0 \\
Task 15 & TurnOnElectricKettle & 3 / 3 & 20/20 & 100.0 \\
Task 16 & TurnOnMicrowave & 2 / 2 & 0/20 & 0.0 \\
Task 17 & TurnOnSinkFaucet & 2 / 2 & 10/20 & 50.0 \\
\midrule
\multicolumn{3}{l}{\textbf{Total}} & 234/360 & \textbf{65.0} \\
\bottomrule
\end{tabular}
\end{table}

\subsection{Harness Zero-shot}
\label{app:harness-zeroshot}

Table~\ref{tab:harness-zeroshot} reports the per-task success rates underlying Fig.~\ref{fig:harness-zeroshot}: the zero-shot transfer of the \CodePhysical{} harness to DreamZero, a World Action Model.
The harness evolved on $\pi_{0.5}$ over LIBERO-Object is applied unchanged to DreamZero, and we report per-task success rate (\%) under the Swap and Task perturbations of LIBERO-PRO.

\begin{table}[htbp]
\centering
\caption{Per-task results of DreamZero and DreamZero with \CodePhysical on LIBERO-PRO Object Suite under Swap and Task perturbation (success rate, \%)}
\label{tab:harness-zeroshot}
\begin{tabular}{lcccc}
\toprule
 & \multicolumn{2}{c}{DreamZero (WAM)} & \multicolumn{2}{c}{\CodePhysical{} (\emph{zero-shot})} \\
\cmidrule(lr){2-3}\cmidrule(lr){4-5}
Task & Swap & Task & Swap & Task \\
\midrule
task0 (alphabet soup) & 0.0 & 0.0 & 100.0 & 94.0 \\
task1 (cream cheese) & 0.0 & 96.0 & 84.0 & 100.0 \\
task2 (salad dressing) & 0.0 & 0.0 & 100.0 & 100.0 \\
task3 (BBQ sauce) & 0.0 & 0.0 & 84.0 & 96.0 \\
task4 (ketchup) & 0.0 & 0.0 & 96.0 & 78.0 \\
task5 (tomato sauce) & 0.0 & 0.0 & 56.0 & 100.0 \\
task6 (butter) & 6.0 & 0.0 & 10.0 & 66.0 \\
task7 (milk) & 0.0 & 0.0 & 46.0 & 34.0 \\
task8 (chocolate pudding) & 0.0 & 0.0 & 98.0 & 96.0 \\
task9 (orange juice) & 0.0 & 0.0 & 60.0 & 80.0 \\
\midrule
Overall & 0.6 & 9.6 & \cellcolor{miBest}\textbf{73.4} & \cellcolor{miBest}\textbf{84.4} \\
\bottomrule
\end{tabular}
\end{table}

\section{Skills Case}
\label{app:skills-case}

The closed loop leaves behind more than working code: each failure is distilled into a problem-repair note mounted on the task-graph node where it occurred (Section~\ref{sec:skills}).
This appendix reproduces one representative place-phase skill, \texttt{align-object-to-basket-mouth}, verbatim as it is stored in the skill library.

\begin{promptfile}{align-object-to-basket-mouth.md}
---
name: align-object-to-basket-mouth
description: Diagnose and repair place-phase 3D alignment in LIBERO-object pick-and-place. Use when the object is 3D-aligned to the basket body instead of the mouth -- the place target is biased toward the camera in depth, the rim Z undershoots the opening, or the object clips the rim / misses the contain_region despite correct 3D object-center alignment. This bias masquerades as a transport drop or a release failure.
phase: place
source_tasks: ["pick-and-place-in-basket"]
---

# align-object-to-basket-mouth

## Problem Encountered
The place phase 3D-servos the object center to `basket + object_to_eef` (see
`align-object-3d-over-basket`). That only drops the object through the opening
when `basket` is the **mouth**. `basket_center_world` returned the median XY of
**all** back-projected mask points plus an 82nd-percentile Z. From the angled
agentview, the visible **near wall** folds into that XY median, biasing the
target toward the camera, and the percentile Z undershoots the true rim.

Measured on `libero_object_task` task2 ep51: the body estimate was
`[0.0585, 0.2653, 0.1321]` while the mouth is `[0.0056, 0.2708, 0.1386]` -- a
**5.3 cm X (depth) error**. Place "succeeded" (the object center reached its
target), yet the object was 5 cm off the opening, clipped the rim, and never
entered the `contain_region`. This was misdiagnosed as a transport drop
(ep51/ep53) and a release timing bug (ep54); all three were one place-target
bias.

## How to Repair
### Workflow
1. Make the place 3D-align target the basket **mouth**. In
   `basket_center_world`, keep only the upper-Z rim points -- the same top-slice
   `target_grasp_world` uses for the object:
   `rim = points[points[:, 2] >= percentile(points[:, 2], BASKET_RIM_QUANTILE)]`.
2. Set the target XY to the **rim median** (mouth center, unbiased in depth) and
   Z to the **rim median** (mouth height). Fall back to all points if fewer than
   5 rim points survive.
3. Leave the place servo (`PlaceMixin.place`) unchanged -- it already aligns the
   object center to `basket + object_to_eef`. With `basket` now the mouth, the
   object drops straight through the opening in one 3D move, no VLA.

### Guardrails
- Do NOT take the basket XY from the all-points median -- the near wall biases it
  toward the camera in depth (the 5 cm X error above). This is the failure mode
  this skill exists to repair.
- Do NOT pair a body-centroid XY with a rim-percentile Z -- compute both from the
  same top-rim point set, or the target is inconsistent in Z vs XY.
- Do NOT chase the single highest-Z pixel (raise `BASKET_RIM_QUANTILE` toward
  the max) -- the median of the top slice is steadier than the max; keep it
  symmetric with `TARGET_TOP_QUANTILE`.
- This skill owns the place **target** (the mouth); pair it with
  `align-object-3d-over-basket` (the object-center servo) and
  `record-3d-grasp-offset` (the 3D EEF->object offset). Do not fold it into
  either -- the target is a distinct failure surface.

## Code Links
`basket_center_world` in `grounding.py` (the target the place servo consumes) of
the task package, e.g.
`capx/eva/code_library/libero/object/task2/grounding.py`. The place servo itself
(`PlaceMixin.place` in `place.py`) needs no change -- it already aligns the object
center to `basket + object_to_eef`; it only succeeds once `basket` is the mouth.

Read `references/basket-mouth.md` for the angled-camera geometry, the measured
bias, and why the fix mirrors `target_grasp_world`.

- `basket_center_world`
- `grounding.py`
- [grounding.py](code_library/libero/object/task2/grounding.py)
- `PlaceMixin.place`
- `place.py`
- `basket + object_to_eef`
- `basket`
- `references/basket-mouth.md`
- `target_grasp_world`
\end{promptfile}